\documentclass[lettersize,journal]{IEEEtran}
\usepackage{amsmath,amsfonts}
\usepackage{threeparttable}
\usepackage{algorithmic}
\usepackage{algorithm}
\usepackage{array}
\newcolumntype{P}[1]{>{\centering\arraybackslash}p{#1}}
\usepackage[caption=false,font=normalsize,labelfont=sf,textfont=sf]{subfig}
\usepackage{textcomp}
\usepackage{stfloats}
\usepackage{url}
\usepackage{verbatim}
\usepackage{graphicx}
\usepackage{cite}
\newtheorem{defka}{Definition}
\newtheorem{thm}{Theorem}

\newtheorem{lem}{Lemma}
\newtheorem{rmk}{Remark}
\newtheorem{prop}{Property}
\newtheorem{pf}{Proof}

\begin{document}

© 2024 IEEE. Personal use of this material is permitted.
Permission from IEEE must be obtained for all other uses,
including reprinting/republishing this material for advertising
or promotional purposes, collecting new collected works
for resale or redistribution to servers or lists, or reuse of
any copyrighted component of this work in other works.
This work has been submitted to the IEEE for possible
publication. Copyright may be transferred without notice,
after which this version may no longer be accessible.

\title{Impact-Resilient Orchestrated Robust Controller for Heavy-duty Hydraulic Manipulators}

\author{Mahdi Hejrati and Jouni Mattila
        % <-this % stops a space
\thanks{This work is supported by Business Finland partnership project "Future all-electric rough terrain autonomous mobile manipulators" (Grant 2334/31/222). Corresponding author: Mahdi Hejrati}% <-this % stops a space
\thanks{Authors are with Department of Engineering and Natural Science, Tampere University,  7320 Tampere, Finland (e-mail: mahdi.hejrati@tuni.fi, jouni.mattila@tuni.fi)}}
% \thanks{M. Hejrati, corresponding author,  is with the Department of Engineering and Natural Science, Tampere University, 7320 Tampere, Finland (e-mail: mahdi.hejrati@tuni.fi).}
% \thanks{J. Mattila is with  the Department of Engineering and Natural Science, Tampere University, 7320 Tampere, Finland (e-mail: jouni.mattila@tuni.fi).}

% The paper headers
\markboth{IEEE Transactions, April 2024}%
{Shell \MakeLowercase{\textit{et al.}}:}

% \IEEEpubid{0000--0000/00\$00.00~\copyright~2021 IEEE}
% Remember, if you use this you must call \IEEEpubidadjcol in the second
% column for its text to clear the IEEEpubid mark.

\maketitle

\begin{abstract}
Heavy-duty operations, typically performed using heavy-duty hydraulic manipulators (HHMs), are susceptible to environmental contact due to tracking errors or sudden environmental changes. Therefore, beyond precise control design, it is crucial that the manipulator be resilient to potential impacts without relying on contact-force sensors, which mostly cannot be utilized. This paper proposes a novel force-sensorless robust impact-resilient controller for a generic 6-degree-of-freedom (DoF) HHM constituting from anthropomorphic arm and spherical wrist mechanisms. The scheme consists of a neuroadaptive subsystem-based impedance controller, which is designed to ensure both accurate tracking of position and orientation with stabilization of HHMs upon contact, along with a novel generalized momentum observer, which is for the first time introduced in Plücker coordinate, to estimate the impact force. Finally, by leveraging the concepts of virtual stability and virtual power flow, the semi-global uniformly ultimately boundedness of the entire system is assured. To demonstrate the efficacy and versatility of the proposed method, extensive experiments were conducted using a generic 6-DoF industrial HHM. The experimental results confirm the exceptional performance of the designed method by achieving a subcentimeter tracking accuracy and by \(80\%\) reduction of impact of the contact.
% The baseline control approach is virtual decomposition control (VDC) method, which is Newton-Euler-based scheme represented in Plücker coordinate with 6D vectors.   
\end{abstract}
\begin{IEEEkeywords}
Impedance control, adaptive neural network, contact force estimation, hydraulic systems.

\end{IEEEkeywords}

\section{Introduction}

\IEEEPARstart{H}{ydraulically} driven manipulators are extensively used in industrial applications such as off-road mobile machines and heavy-duty operations due to their distinct advantages, including a favorable size-to-power ratio and the ability to generate substantial output force or torque compared to their electrical counterparts \cite{mattila2017survey,cheng2023prioritized,yao2013high, jud2021heap}. However, despite these benefits, design of a stable, high-precision controller for hydraulic actuators is challenging due to the governing nonlinear fluid dynamics, various input constraints (e.g., backlash, deadzone, and saturation), and the complexities of unmodeled uncertainties and closed-chain mechanisms, particularly as the number of degrees of freedom (DoF) increases \cite{xu2022extended,mattila2017survey}. Thus, building a high-performance controller with subcentimeter accuracy is critically important for the autonomous operation of heavy-duty hydraulic manipulators (HHMs). 

Numerous studies have addressed the control challenges of hydraulically actuated manipulators in free-motion tasks. Various approaches have been employed for joint motion control, including adaptive control \cite{xu2022extended}, adaptive robust control \cite{mohanty2010integrated}, neuro-adaptive backstepping sliding mode control \cite{truong2023backstepping}, model predictive control \cite{mononen2019nonlinear}, and intelligent control methods such as reinforcement learning \cite{yao2023data} and radial basis function neural networks (RBFNNs) \cite{liang2024adaptive}. Another widely used approach is virtual decomposition control (VDC) \cite{zhu2010virtual} for the motion control of hydraulic manipulators \cite{koivumaki2015high,koivumaki2019energy}. The focus on free-motion control is due to the prevalence of optimization-based, collision-free trajectory generation in path planning for autonomous operations within known or partially known environments \cite{zhou2020robust,preiss2017trajectory}. However, in unstructured environments, collision avoidance is challenging due to the lack of precise environmental knowledge. Even in fully known environments, tracking errors in the system can lead to unintended contact, which can cause irreparable damage to both the robot and the environment, especially in heavy-duty operation with huge interaction forces. Therefore, designing impact-resilient controllers or developing collision-inclusive motion planning to reduce the contact impact in contact-rich operations is crucial and has garnered significant research attention \cite{lu2022online,lu2021deformation}. This paper aims to address these issues in the context of HHMs by designing an impact-resilient control scheme.

Despite the industrial significance of HHMs, limited studies have focused on designing controllers for contact-rich manipulation tasks. In \cite{ha2000impedance}, a hybrid force/motion controller was developed for a backhoe excavator using the sliding mode approach. To address the contact problem during the shoveling phase of hydraulic mining machines, \cite{qin2022adaptive} proposed an adaptive, robust impedance controller. A VDC-based hybrid motion/force controller was designed in \cite{koivumaki2015stability} for a 2-DoF HHM. An adaptive impedance controller for contact force tracking in hydraulic excavators was proposed in \cite{feng2022adaptive}. A stability-guaranteed impedance controller was designed in \cite{koivumaki2016stability} to ensure the performance of the VDC scheme in the event of unexpected contacts. The benefits of a backdrivable servovalve for impedance control were thoroughly analyzed and experimentally verified in \cite{yoo2019impedance}. The problem of contact with uncertain environments was addressed in \cite{ding2023adaptive} through the design of an adaptive impedance controller. Although these studies have conducted experiments to validate their results, they typically focused on manipulators with fewer than three DoFs, simplifying the implementation. However, in real-world applications, enhancing the dexterity of manipulators often requires a higher number of DoFs, which increases system complexity and complicates the implementation of the designed controllers. Thus, there is a critical need for a high-precision, impact-resilient controller that can be feasibly implemented on high-DoF HHMs in industrial applications. In this study, by extending our previous research \cite{hejrati2023orchestrated}, which only considered free-motion tasks, a VDC-based, impact-resilient controller is designed to address these issues.

A key challenge in impact-resilient control design for HHMs is measuring contact force. Placing 6-DoF force/torque sensors on the end-effector of HHMs is not always feasible due to factors such as high costs, potential damage to the sensors during operations, and the complexity of their retrofitting. Consequently, alternative methods for estimating or measuring contact force need to be explored to ensure effective and reliable impact-resilient control. One of the well-known approaches for external force/torque estimation is generalized momentum observer (GMO) \cite{de2006collision,yoo2019impedance}, which requires a Lagrangian representation of system dynamics. As the baseline controller in this study, VDC scheme uses Newton-Euler (NE) approach to represent the dynamics of the system in Plücker coordinates. Therefore, direct integration of GMO and VDC could result in computational inefficiencies and complicate real-time implementation due to the necessity of dual model representations. Given the exceptional performance of the VDC in HHM applications \cite{koivumaki2019energy,koivumaki2015high,hejrati2023orchestrated} and its widespread use in diverse fields \cite{lampinen2021force, ding2022vdc, hejrati2023nonlinear}, addressing the aforementioned challenge becomes imperative. Existing solutions, such as the gravity-compensation-based estimator \cite{koivumaki2015stability} circumvent some of these challenges by ignoring inertial and centrifugal effects, which is effective at low speeds. Additionally it requires Lagrangian-based gravity vector derivations, which is incompatible with the VDC context. Recognizing the systematic advantages of VDC in modeling, control, and stability analysis for complex systems, this paper proposes a novel GMO-based force estimator introduced in Plücker coordinates to tackle the issues described above.

Considering the aforementioned issues, the current work designs a novel force-sensorless impact-resilient controller for generic 6-DoF real-world HHM with anthropomorphic arm and spherical wrist. The proposed high-level impact-resilient command is executed by joint-level, robust neuro-adaptive controller, designed in our previous work \cite{hejrati2023orchestrated}. Moreover, new formulation for GMO is introduced in the Plücker coordinate that refines the GMO and VDC integration, making it suitable for real-world implementation. Moreover, the robustness and stability of the system under designed controller are proved by means of virtual stability and virtual power flow (VPF) concepts. Consequently, the contributions of the current work can be expressed as follows:
\begin{itemize}
    \item A force-sensorless impact-resilient method is proposed for full-pose control of a generic 6-DoF industrial HHM with an anthropomorphic arm and spherical wrist, demonstrating the method's general applicability.
    \item For the first time, a GMO-based force estimator is developed in Plücker coordinates and embedded into VDC.
    \item Extensive real-time experiments on full-scale 6-DoF industrial HHM is conducted to show the performance and universality of the proposed scheme.
\end{itemize}

The rest of the paper is organized as follows. Section II expresses the fundamental mathematics of 6D vectors. Section III describes the impact force estimation developed in Plücker coordinate, whereas in Section IV, the details of the proposed impact-resilient controller and stability analysis are expressed. The experimental results are provided in Section V for performance evaluation. Finally, Section VI concludes this study.
% By building upon our previous work with neuro-adaptive, high-precision joint controller, we propose an impact-resilient, adaptive impedance controller for position and orientation control of 6-DoF HHM. In order to remove the extra computational burden of exploying non-VDC-exclusive force estimation algorithms and making it suitable for real-time implementation, an embedded, subsystem-based force estimator is proposed. The mentioned two contributions make the VDC a suitable approach to be implemented on HHMs in industrial application.

\begin{figure}
    \centering
    \includegraphics[width=0.4\textwidth]{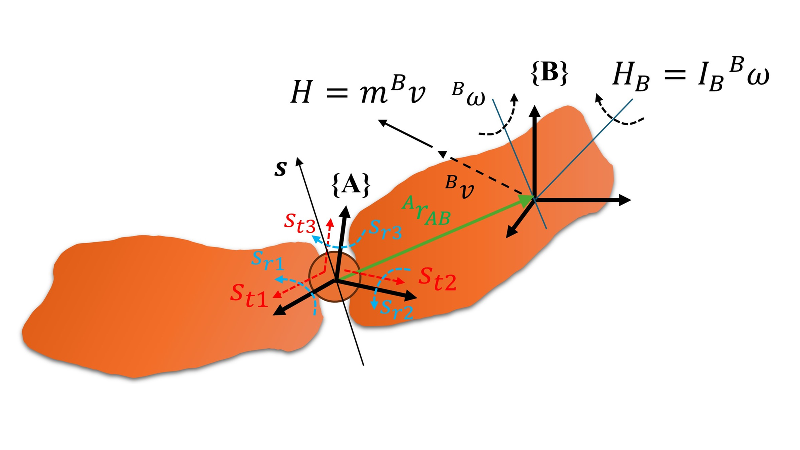}
    \caption{Interconnected rigid bodies with attached Plücker coordinates.}
    \label{Fig bodies}
\end{figure}

\section{Mathematical Formulation and Foundations}

\subsection{Spatial Force and Velocity Vectors in Plücker Coordinates}
VDC formulation is based on Plücker coordinate with 6D Plücker basis which results in a compact formulations of the system dynamics \cite{featherstone2014rigid}. Consider \(\mathcal{M}^6\) as the spatial motion vector space, and its dual space \(\mathcal{F}^6\), as the spatial force vector space. Consequently, by considering \{A\} as a frame that is attached to a rigid body (Fig. \ref{Fig bodies}), the spatial velocity vector \(^A\mathcal{V}\in \mathcal{M}^6\) and spatial force vector \(^A{F}\in \mathcal{F}^6\) can be expressed as follows \cite{zhu2010virtual}:
\begin{equation*}
^A\mathcal{V} = [^Av,\,^A\omega]^T,\quad ^AF = [\,^Af,\,^Am]^T
\end{equation*}
with \(^Av\) and \(^A\omega\) being two 3D coordinate vectors as linear and angular velocities, respectively, and \(^Af\) and \(^Am\) being 3D coordinate vectors representing linear force and moment, respectively, of frame \{A\}. It is assumed that each joint only allows a single degree of motion freedom, so that the linear/angular velocity vector can be described as:
\begin{equation} \label{V=sq}
^A\mathcal{V} = \boldsymbol{s}\,\Dot{q}
\end{equation}
with \(\Dot{q} \in \Re\) being joint velocity and \(\boldsymbol{s} = [\boldsymbol{s}_t^T,\, \boldsymbol{s}_r^T]^T\) being unit vector that determines the axis of the joint motion (as depicted in Fig. \ref{Fig bodies}), with \(\boldsymbol{s}_t\) being translation vector and \(\boldsymbol{s}_r\) being rotation vector. The transformation matrix that transforms spatial force and velocity vectors between frames \{A\} and \{B\} (which coincides with the center of mass of the rigid body) is \cite{zhu2010virtual},

\begin{equation}\label{equ1}
^AU_B = \begin{bmatrix}
^AR_B & \textbf{0}_{3\times3} \\
(^Ar_{AB}\times)\, ^AR_B & ^AR_B
\end{bmatrix}
\end{equation}
where \(^AR_B \in \Re^{3\times3} \) is a rotation matrix between frame \{A\} and \{B\}, and \(^Ar_{AB} = [r_x,r_y,r_z]^T\) denotes a vector from the origin of frame \{A\} to the origin of frame \{B\}, expressed in \{A\} with \((\times\)) operator defined in Appendix A. Consequently, the spatial force and velocity vectors can be transformed between frames, as \cite{zhu2010virtual},
% \begin{equation}\label{equ2}
% ^Ar_{AB}\times = \begin{bmatrix}
% 0 & -r_z & r_y \\
% r_z & 0 & -r_x\\
% -r_y & r_x & 0
% \end{bmatrix}
% \end{equation}
% where . If \(^AU_B\) performs a coordinate transform on force vector, then \(^AU_B^T\) performs the same transform on its dual vector that is motion vector. 
\begin{equation}\label{equ3}
^B\mathcal{V} =\, ^AU_B^T\,^A\mathcal{V},\quad ^BF =\, ^AU_B\, ^BF.
\end{equation}
\newpage
\subsection{Momentum}
For the given rigid body configuration in Fig. \ref{Fig bodies} with attached frames of \{A\} and \{B\} and given spatial velocity \(^B\mathcal{V} = [^Bv,\,^B\omega]^T\) with inertial parameters \(m\) and \(I_B\) as mass and rotational inertia, respectively, the linear and angular momentum can be defined as \(H = m\,^Bv\) and \(H_B = I_B\,^B\omega\), respectively. As the spatial momentum vectors \(\Bar{H}_B = [H^T,\,H_B^T]^T\) are elements of \(\mathcal{F}^6\), they have the same algebraic properties as other spatial force vectors. Therefore, the spatial momentum vector of frame \{A\} can be described as,
\begin{equation}\label{hA = Uhb}
    H_A =\, ^AU_B\,\Bar{H}_B =\, (^AU_B\,M_B\,^AU_B^T)\,^AV = M_A\,^AV
\end{equation}
with \(M_B = diag([m\,\cdot I_3, I_B)]\), and \(M_A\) can be derived by performing matrix multiplication:
\begin{equation} \label{M_A}
M_A = \begin{bmatrix}
m_A\, I_3 & -m_A\, (^Ar_{AB}\times) \\
m_A\, (^Ar_{AB}\times) & I_A
\end{bmatrix}\\,
\end{equation}
with \(I_A =\, ^AR_B\,M_B\,^AR_B^T-m(^Ar_{AB}\times)^2\). Equation (\ref{M_A}) is the general expression for spatial inertia matrix in Plücker coordinates.

\begin{prop}
     Since \(I_B\) is symmetric matrix, then both \(M_B\) and \(M_A\) are symmetric matrix as well. Additionally, if \(m>0\) and \(I_B\) is positive definite, then both \(M_B\) and \(M_A\) are positive definite.
    \label{property 1}
\end{prop}

\subsection{Rigid Body Dynamics}
The dynamic equation of the rigid body in space, shown in Fig. \ref{Fig bodies}, whic is based on unique inertial parameter vector function \(\phi(m,\,^Ar_{AB}, I_A)  \in \Re^{10}\) expressed in frame \{A\}, can be written as\cite{hejrati2022decentralized}:

\begin{equation}\label{F*}
M_A\frac{d}{dt}(^A\mathcal{V})+C_A (^{A}\omega)\,{^A\mathcal{V}}+G_A + \Delta_A=\, ^AF^*
\end{equation}
where \(M_A \in \Re^{6\times6} \) is the mass matrix, \(C_A \in \Re^{6\times6} \) is the centrifugal and Coriolis matrix, \(G_A \in \Re^6 \) is the gravity vector, \(\Delta_A \in \mathcal{F}^6\) is uncertainty stemming from the rigid body model, and \(^AF^* \in \mathcal{F}^6 \) is the net spatial force vector applied to the rigid body. 

\begin{prop}
    For a given rigid body dynamics (\ref{F*}), the following property holds:
    \begin{equation*}
        \Bar{Y}_{A} \phi_{A} = M_{A} \dfrac{\rm d}{\mathrm{d}t} \left( {^{A}V} \right) + C_{A} \left( {^{A}{\omega}}  \right) {^{A}V} + G_{A}.
    \end{equation*}
    with \(\Bar{Y}_{A} \in \Re^{6\times10}\) being regression matrix, and \(\phi_A(m,\,^Ar_{AB}, I_A)\) being unique inertial parameter vector \cite{hejrati2022decentralized}.
    \label{property 2}
\end{prop}

In the VDC approach, the required velocity plays an important role by encompassing the desired velocity along with one or two error terms related to the position or force error in the position or complaint control mode, respectively. Considering joint space required velocity as \(\Dot{q}_r\), which will be defined based on a given task, and computing the spatial velocities of the connected rigid bodies and exploiting property \ref{property 2}, the required net force/moment vector can be defined as \cite{hejrati2022decentralized},
\begin{equation}\label{F*r}
^AF_r^* = Y_{A} \hat{\phi}_{A} +\,^AF_c
\end{equation}
where \(^AF_c\) is the feedback control term, and \(Y_{A} \hat{\phi}_{A}\) is feed forward term, where \(\hat{\phi}_A\) is the estimation of \(\phi_A\).
\newpage
% \begin{rmk}
%     By computing (\ref{F*}) and (\ref{F*r}) for each rigid body at the corresponding joint, the actuator forces \(f_c = \boldsymbol{s}^T\,^AF^*\) and \(f_{cr}= \boldsymbol{s}^T\,^AF_r^*\) can be computed, representing the force that must be generated at the actuator level to accomplish the rigid body and joint level control goals.
%     \label{fc remark}
% \end{rmk}

\begin{figure}[t]
      \centering
      \subfloat[]{\includegraphics[width = 0.35\textwidth]{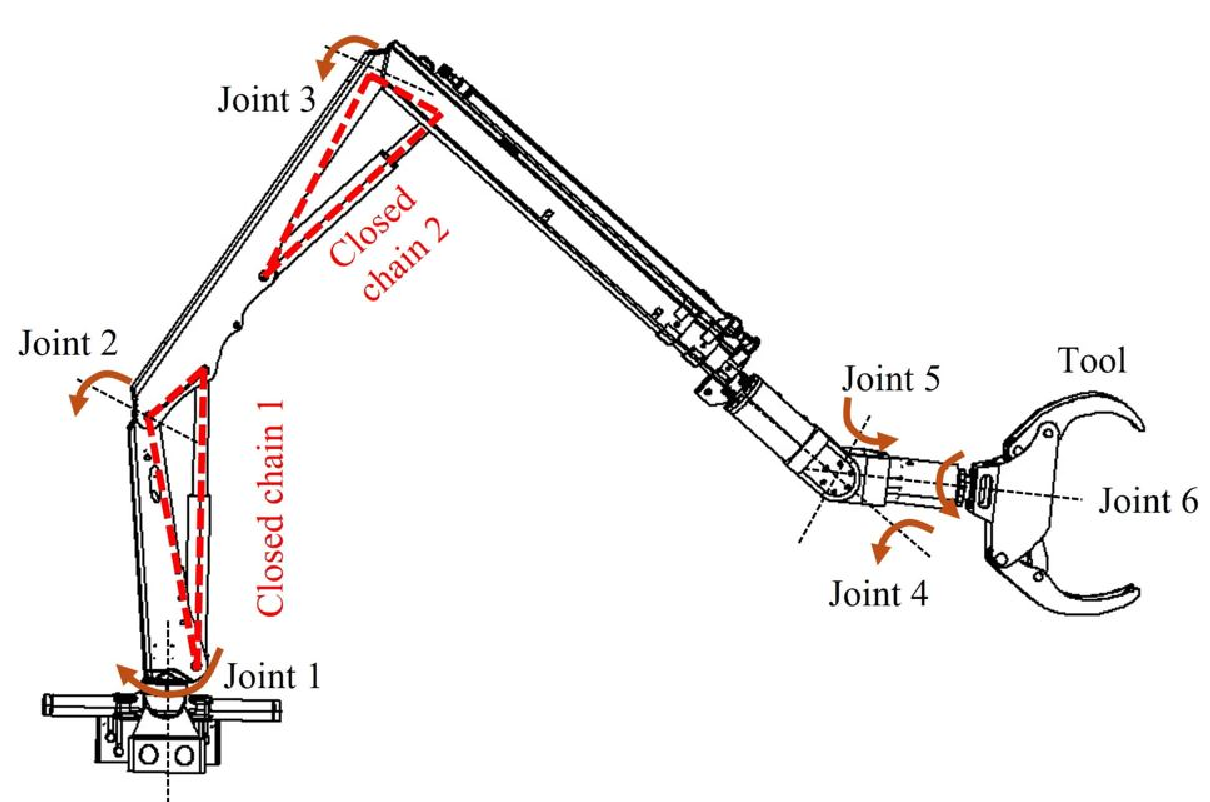}
      \centering
      \label{fig4}}
      \hfil
      \subfloat[]{\includegraphics[width = 0.45\textwidth]{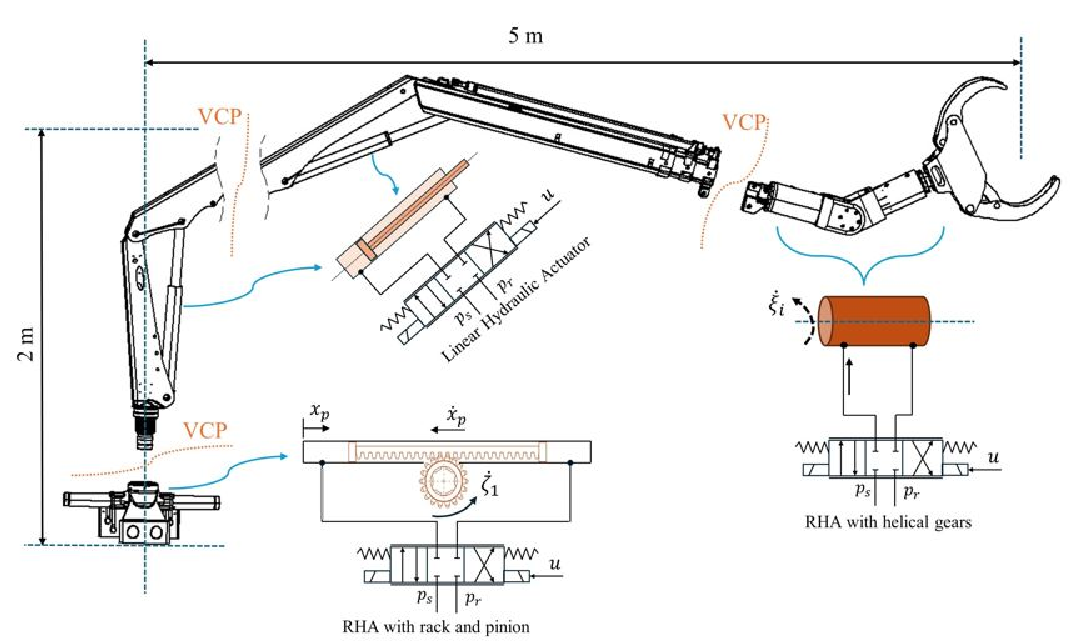}
      \centering
      \label{fig5}}
      \hfil
      \subfloat[]{\includegraphics[width = 0.45\textwidth]{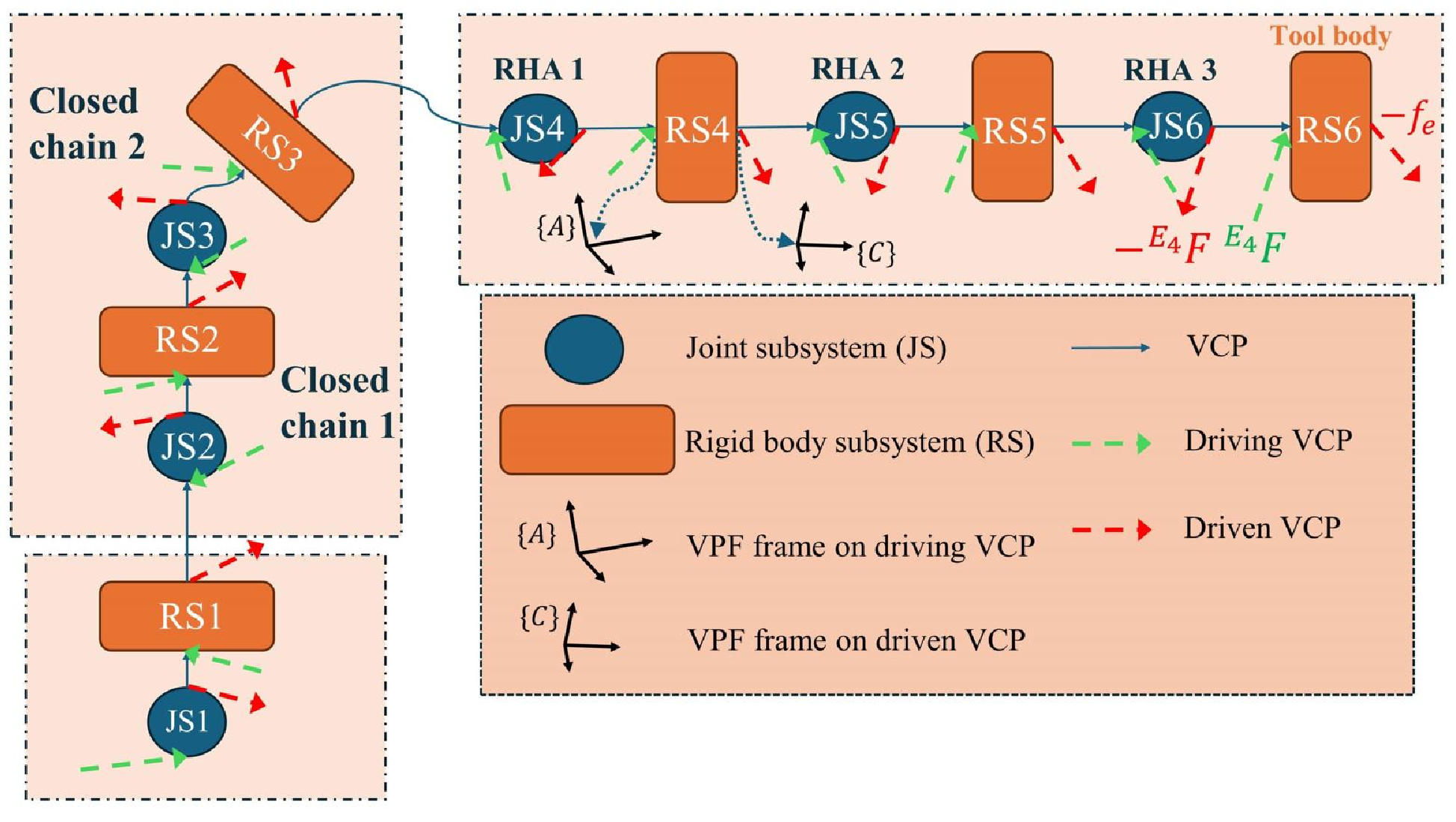}
      \centering
      \label{fig6}}
      \caption{Scheme of 6-DoF heavy-duty hydraulic manipulator (HHM) with VDC approach. a) general configuration of HHM, b) decomposition of HHM into subsystems, c) Scheme of VDC with VCP concept}
      \label{HIAB}
   \end{figure}

\subsection{Virtual Stability}

The VDC approach divides a complex system into subsystems using virtual cutting points (VCPs), as illustrated in Fig. \ref{HIAB}. A VCP forms a virtual cutting surface (Fig. \ref{fig5}) where spatial force vectors can be exerted from one part, which is interpreted as a driving VCP, to another, which is interpreted as a driven VCP, shown with green and red lines in Fig. \ref{fig6}, respectively. Once the system is decomposed by placing VCPs, control actions can be designed to meet subsystem-level objectives. Another key concept in VDC is virtual power flow (VPF), which characterizes dynamic interactions among subsystems. Presented in Deﬁnition \ref{Def pA}, the VPF at a driving cutting point has the same magnitude as the VPF at the corresponding driven point, but the opposite sign. Thus, the sum of VPFs at all driving points equals the sum at all driven points. In the followings, important definitions in the VDC context is provided.
% The VDC approach breaks down the entire complex system into subsystems by means of a virtual cutting point (VCP), as shown in Fig. \ref{HIAB}. A VCP forms a virtual cutting surface (displayed in Fig. \ref{fig5}) on which spatial force vectors can be exerted from one part to another, as demonstrated with green and red lines in Fig. \ref{fig6}. The cutting point is interpreted as a driving cutting point by one part and is simultaneously interpreted as a driven cutting point by another part. The spatial force vector is exerted from one part, where the cutting point is interpreted as a driving cutting point, to the other part, where the cutting point is interpreted as a driven cutting point. After the original system is virtually decomposed into subsystems by placing VCPs, the control action can be designed to achieve the subsystem-level objectives. Another term in the context of VDC is virtual power flow (VPF), that is used to characterize the dynamic interactions among subsystems. Being a scalar, the VPF by Deﬁnition \ref{Def pA} at a driving cutting point takes exactly the same magnitude as the VPF at the corresponding driven cutting point, but the sign. Therefore, the sum of VPFs at all driving cutting points equals the sum of VPFs at all driven cutting points. In the following the important definition of VDC context are provided.
\begin{defka}
    With respect to frame \{A\}, the VPF is deﬁned as the inner product of the spatial velocity vector error and the spatial force vector error, that is,
    \begin{equation*}\label{VPF}
    p_A = (^A\mathcal{V}_r-\,^A\mathcal{V})^T(^AF_r-\,^AF)
    \end{equation*}
    where \(^A\mathcal{V}_r \in \mathcal{M}^6\) and \(^AF_r \in \mathcal{F}^6\) represent the required vectors of \(^A\mathcal{V} \in \mathcal{M}^6\) and \(^AF \in \mathcal{F}^6\), respectively. The inner product indicates the duality relationship between \(\mathcal{M}^6\) and \(\mathcal{F}^6\).
    \label{Def pA}
\end{defka}
\vspace{-4mm}
\begin{defka}
    A non-negative accompanying function \(\nu(t) \in \Re\) is a piecewise, differentiable function defined \(\forall \, t \in \Re^+\) with \(\nu(0) < \infty\) and \(\Dot{\nu}(t)\) exists almost everywhere.
    \label{nu def}
\end{defka}
\begin{defka}
    \cite{hejrati2023orchestrated} Consider a complex robot that is virtually decomposed into subsystems. Each subsystem is said to be virtually semi-globally uniformly ultimately bounded with its non-negative accompanying function \(\nu(t)\) and its affiliated vector \(\Dot{\nu}(t)\), if and only if,
\begin{equation}\label{equ9}
\Dot{{\nu}} \leq -\alpha_1 {\nu}_1 + \alpha_{10}+p_A-p_C
\end{equation}
with \(\alpha_1\) and \(\alpha_0\) being positive and \(p_A\) and \(p_C\) denoting the sum of VPFs in the sense of Definition 2 at frames \{A\} (placed at driving VCPs) and \{C\} (placed at driven VCPs), shown in Fig. \ref{fig6}.
\end{defka}
\begin{thm}
    Consider a complex robot (Fig. \ref{fig4}) that is virtually decomposed into subsystems (Fig. \ref{fig6}). If all the decomposed subsystems are virtually semi-globally uniformly ultimately bounded in the sense of Definition 3, then the entire system has semi-global uniformly ultimate boundedness (SGUUB)\cite{zhu2010virtual,hejrati2023orchestrated} .
    \label{thm1}
\end{thm}

\begin{rmk}
    Theorem 1 is the most important theorem in the VDC context. It establishes the equivalence between the virtual stability of every subsystem and the stability of the entire complex robot. 
\end{rmk}

\section{ Impact Force Estimation}

% \subsection{Environment Modeling}
% In general, every environment possesses a certain amount of ﬂexibility. To determine an environment as ﬂexible or rigid is completely dependent on its mechanical impedance. If an environment possesses a comparable or lower mechanical impedance than that of a controlled robot, it is considered a ﬂexible environment. If an environment possesses a much higher mechanical impedance than that of a controlled robot, it is considered a rigid environment. In this paper, both of the environments are taken into account.
% Fig. \ref{Wrist_contact} depicts the end-effector of the robot approaching an unknown environment. Frame $\{E_4\}$ is attached to the tool of the robot, frame $\{E\}$ is fixed to the end-effector, and frame $\{T\}$ is the contact point of the robot tool with the environment.
% The governing dynamics of the rigid environment can be expressed as,
% \begin{equation}
%     \Dot{X}_{er} = (1-\sigma)V_T
% \end{equation}
% \begin{equation}
%     f_{er} = \sigma\,f_e
% \end{equation}
At the time of impact, the environment shows impedance behavior, outputting force for the given input velocity. Such a behaviour can be modeled as,
\begin{equation} \label{Env model}
     M_f\, \Ddot{X}_{ef} + D_f\,\Dot{X}_{ef} + K_f\, \Tilde{X}_{ef}  = f_e
\end{equation}
with \(M_f\), \(D_f\), \(K_f\) being inertia, damping, and stiffness of environment, \(\Tilde{X}_{ef} \) denoting the deformation of the environment, while \(\Dot{X}_{ef}\) and \(\Ddot{X}_{ef}\) are equal to the end-effector velocity and acceleration.  From (\ref{Env model}) can be concluded that by having the position and parameters of the environment, the contact force can be computed. However, in most of the real-world implementations neither of the information are available, requiring direct force estimation. In this section, a novel GMO-based force estimator is presented in Plücker coordinate for impact force estimation.

\subsection{Generalized momentum observer}
% At the time that contact happens, the environment plays the impedance role and the manipulator the admittance role; in this way, the velocity out of the robot goes as the input to the environment model and results in contact force. Therefore, it is important to identify the parameters of the environment and estimate the contact force.
The current form of GMO for contact force estimation requires the Lagrangian representation of the robot dynamics as,
\begin{equation} \label{Lag_model}
    M\,\Ddot{x} + C(x,\Dot{x})\,\Dot{x} + G(x) = \tau + J^T\,f_e.
\end{equation}
\newpage
with \(M\), \(C\),\(G\) being mass, centerfugal, and gravity matrices, \(\tau\) is input torque, \(f_e\) being contact force, and \(J\) being Jacobian matrix. The \(C\) matrix must be derived in a way that the property \(\Dot{M}-2C = 0\) holds. Consequently, incorporating GMO into VDC would require the derivation of both Lagrangian model (\ref{Lag_model}) (for GMO design) and NE equations in Plücker coordinate (\ref{F*}) (for control design). In addition to the mathematical burden of deriving two dynamics, the computational cost of such an approach might engender problems in real-time implementation, especially for high-DoF HHMs. In this paper, for the first time, we developed GMO in Plücker coordinate that matches the motion equations of VDC, addressing all the mentioned issues.

Consider a rigid body in space (Fig. \ref{Fig bodies}) represented in Plücker coordinate with spatial velocity and force vectors belonging to \(\mathcal{M}^6\) and \(\mathcal{F}^6\), respectively. The momentum in the frame \{A\} can be represented as (\ref{hA = Uhb}) with time derivative as,
% \begin{equation} \label{HA}
%     H_A = M_A\,^A\mathcal{V}.
% \end{equation}
% Taking the time derivative of (\ref{HA}), one can obtain,
\begin{equation} \label{DHA}
    \frac{d}{dt}{H_A} = \frac{d}{dt}(M_A)\,^A\mathcal{V}+\,M_A\,\frac{d}{dt}(^A\mathcal{V}).
\end{equation}
As the manipulator system is an interconnected multi rigid body, the contact at the end-effector will impact all the rigid bodies of the system. We denote this impact as \(^AF_d\),
\begin{equation}\label{F*2}
^AF_d = M_A\frac{d}{dt}(^A\mathcal{V})+C_A (^{A}\omega)\,{^A\mathcal{V}}+G_A - \, ^AF^*
\end{equation}
Now, from (\ref{F*2}), we can obtain,
\begin{equation} \label{DVA}
    M_A\frac{d}{dt}(^A\mathcal{V}) = \, ^AF^*+\, ^AF_d - (C_A (^{A}\omega)\,{^A\mathcal{V}}+G_A).
\end{equation}
In order to eliminate the need for time derivative of spatial inertial matrix in (\ref{DHA}), we have,
\begin{equation} \label{DMA}
     \frac{d}{dt}(M_A) = -(^A\mathcal{V}\times)^T\,M_A - M_A\,^A\mathcal{V}\times
\end{equation}
where the proof is provided  in Appendix A. By replacing (\ref{DMA}) and (\ref{DVA}) in (\ref{DHA}), one can obtain:
\begin{equation} \label{DHA2}
\begin{split}
    \frac{d}{dt}{H_A} &= -(^A\mathcal{V}\times)^T\,M_A - M_A(^A\mathcal{V}\times)+\, ^AF^*+\, ^AF_d \\
    &- (C_A (^{A}\omega)\,{^A\mathcal{V}}+G_A).
\end{split}
\end{equation}
The representation of momentum variation in each rigid body, as shown in (\ref{DHA2}), allows us to detect the impact of contact at each subsystem that matches the characteristics of the VDC context. By defining,
\begin{equation} \label{mathF}
\begin{split}
    ^A\mathfrak{F}& = \, ^AF^* - (^A\mathcal{V}\times)^T\,M_A - M_A(^A\mathcal{V}\times)\, \\
    &- (C_A (^{A}\omega)\,{^A\mathcal{V}}+G_A)
\end{split}
\end{equation}
one can rewrite (\ref{DHA2}) as,
\begin{equation}\label{Simp Fd}
    ^AF_d = \frac{d}{dt}{H_A} - ^A\mathfrak{F}.
\end{equation}
Now, the residual vector $\mathcal{R}_A$ can be defined as,
\begin{equation}
    \mathcal{R}_A = \boldsymbol{K}\left\lbrace{H_A(t)-H_A(t_0)-\int_{t0}^{t}(^A\mathfrak{F}+\mathcal{R}_A)dt}\right\rbrace,
    \label{residual}
\end{equation}
where $\mathcal{R}_A(t_0) = \boldsymbol{0}$ and $\boldsymbol{K}$ is diagonal positive definite matrix. By taking the derivative of (\ref{residual}) and using (\ref{Simp Fd}), one can obtain,
\begin{equation} \label{Res}
    \frac{d}{dt}\mathcal{R}_A = -\boldsymbol{K}\,\mathcal{R}_A+\,\boldsymbol{K}\,^AF_d.
\end{equation}
Finally, by taking the Laplace transform of (\ref{Res}), we have,
\begin{equation}\label{laplac}
    \mathfrak{r}_{i}(s) = \frac{\boldsymbol{K}_{ii}}{s_i + \boldsymbol{K}_{ii}}\, \mathfrak{f}_i(s)
\end{equation}
with \(\mathcal{R}_A = [\mathfrak{r}_{1},...,\mathfrak{r}_{6}]^T\) and \(^AF_d = [\mathfrak{f}_{1},...,\mathfrak{f}_{6}]^T\). The equation (\ref{laplac}) demonstrates that by setting a high gain to $\boldsymbol{K}$, one can establish $\mathfrak{r}_{i} \simeq \mathfrak{f}_i$, indicating $\mathcal{R}_A \simeq \,^AF_d$.

The residual \(\mathcal{R}_A\) captures the projection of the impact on a rigid body expressed in attached Plücker coordinate \{A\}. Thus, the accumulation of the projections over all the rigid bodies should be computed, from which the estimation of impact force can be reconstructed. For this aim, the map from Plücker coordinate to joint space, and from joint space to end-effector (where impact occurs) is required. The former one can be accomplished by use of unit screw vector \(\boldsymbol{s}\) in (\ref{V=sq}) as, 
% computes the residual vector of a rigid body in Plücker coordinate \{A\}, which is element of \(\mathcal{F}^6\). As \(^AF_d\) is the impact of contact force on each rigid body of complex system, the estimation of the contact force from these impacts will be their accumulation over entire system. Therefore, we need to consider the motion constraints on the rigid body that resulted by joints at connection point of two rigid bodies. For doing so, we use the screw vector to compute the scalar residual of the rigid body on the non-constraint direction as below,
\begin{equation}\label{scalar r}
    r_A = \boldsymbol{s}^T\,\mathcal{R}_A,
\end{equation}
where the latter one can be established by means of Jacobian matrix,
\begin{equation}
    \hat{f}_e = \left (J^T\right)^\dagger \boldsymbol{r},
    \label{fe_hat}
\end{equation}
with \(A \in \mathfrak{G}\), where \(\mathfrak{G}\) includes \(n\) Plücker frames attached to each joint of interconnected multi-body system with \(n\) being the number of joints or DoF of the entire system, \(\boldsymbol{r} = [r_1,..,r_n]^T\) computed through (\ref{scalar r}), and \((.)^\dagger\) being the psuedo-inverse (if non-square) of the Jacobian matrix \(J \in \Re^{6\times 6}\). It must be mentioned that the impact can be detected whenever \(\|\boldsymbol{r}\|>\Bar{r}\), which \(\Bar{r}>0\) is a suitable threshold to prevent false impact alarm.

\begin{rmk}
    The spatial residual (\ref{residual}) needs the spatial vectors and matrices through (\ref{hA = Uhb}) and (\ref{mathF}) defined in Plücker coordinate. This novel way of developing GMO in Plücker coordinates removes the need for derivation of the Lagrangian model for any complex robotic system.
\end{rmk}

\section{Impact Resilient Control Design and Stability Analysis}
In our latest work \cite{hejrati2023orchestrated}, we conducted a comprehensive analysis of kinematics and dynamics for HHM represented in Fig. \ref{HIAB}, and designed a joint-level control action. Building upon this foundation, our current work extends those findings to develop an impact-resilient control scheme. Specifically, we redefine the required velocity \(\Dot{q}_r\) to meet free-motion and contact-rich tasks objectives, following a similar approach to \cite{koivumaki2016stability}. The new required velocity definition will be integrated into the existing neuro-adaptive, orchestrated robust controller from \cite{hejrati2023orchestrated}. For clarity, we will also provide a brief overview of the control and stability aspects.

% In our latest work \cite{hejrati2023orchestrated}, the comprehensive kinematics and dynamics analysis along with joint-level control action design was conducted for a parallel-serial HHM. In order to design an impact resilient control upon \cite{hejrati2023orchestrated}, it is only required to redefine the required velocity \(\Dot{q}_r\) (the same approach utilized in \cite{koivumaki2016stability}). Therefore, in this paper, we will define required velocity to accomplish the task space objectives, while the new control command will be executed by the same neruo-adaptive-based, orchestrated robust controller in \cite{hejrati2023orchestrated}. However, for readability, a brief elaboration on control and stability will be provided.
\subsection{Impact Resilient Control}

\begin{figure*}
    \centering
    \includegraphics[width=0.9\textwidth]{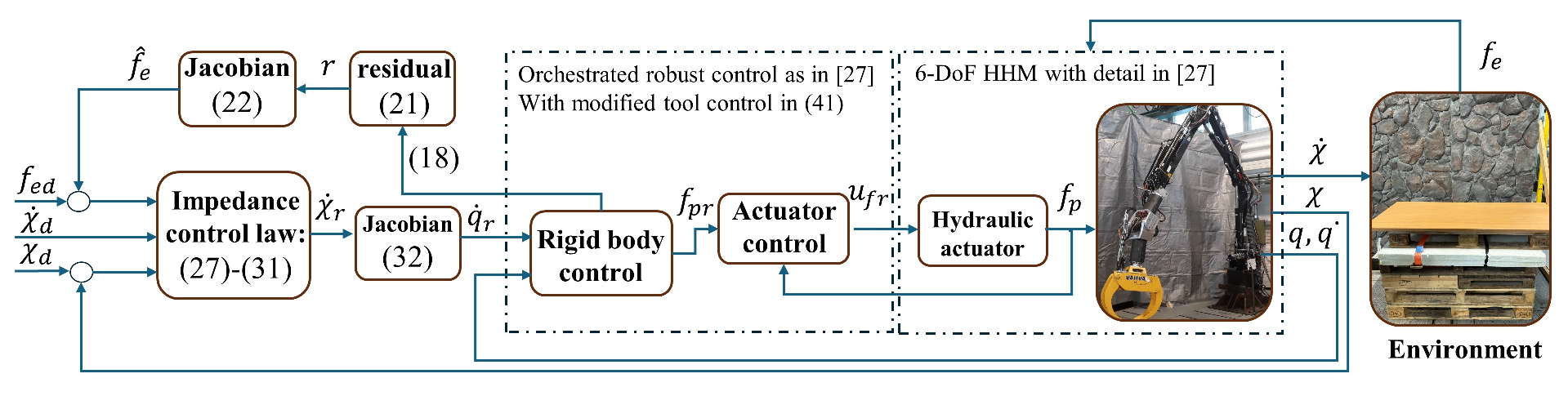}
    \caption{ Impact resilient control scheme. The impedance law \(\dot{\mathcal{X}}_r\) computed in (\ref{DXr}), is transformed into joint space command \(\Dot{q}_r\) which is executed by the control approach in \cite{hejrati2023orchestrated}. The actuator-level control command, \(u_{fr}\) being desired voltage, is designed to generated required piston force \(f_{pr}\), accomplishing control objectives. The command \(u_{fr}\) controls the hydraulic valves. Please see \cite{hejrati2023orchestrated} for detailed explanation.}
    \label{Scheme}
\end{figure*}

% The designed impact resilient approach in this study is VDC-based impedance control. Impedance control, introduced by Hogan \cite{hogan1984impedance}, attempts to control the relationship between force and position of the end-effector in contact with the environment (shown in Fig. \ref{Wrist_contact}), yet ensuring tracking performance in free motion. 
The pose of the end-effector can be written as function of joint angles \(q\),
\begin{equation}
    \mathcal{X} = \mathcal{N}(\boldsymbol{q})
\end{equation}
where \(\mathcal{N}(.)\) shows forward kinematics of the manipulator. Then, the corresponding velocity of end-effector can be computed as,
\begin{equation}
    \Dot{\mathcal{X}} = J(\boldsymbol{q})\dot{\boldsymbol{q}}
\end{equation}
with \(\dot{q}\) being joint velocity vector. Additionally, the force-velocity characteristic at the time of impact, can be written as,
\begin{equation} \label{Z}
    \mathcal{Z}(s) = \dfrac{\mathcal{F}_e(s)}{\Dot{\mathcal{X}}(s)} = M_d\,s + D_d\, + \, K_d\,\dfrac{1}{s}
\end{equation}
with \(M_d\), \(D_d\), \(K_d\) being desired mass, damping, and stiffness matrices, and \(s\) being Laplace transformer. The goal is to design a control law that ensures the desired impedance behaviour for HHMs. For doing so, 
we can rewrite (\ref{Z}) in Cartesian space by omitting inertia as,
\begin{equation}
    D_d\,(\Dot{\mathcal{X}}_d - \Dot{\mathcal{X}}) + K_d\, (\mathcal{X}_d - \mathcal{X}) = -(f_{ed} - f_e),
    \label{des impedance}
\end{equation}
where $\mathcal{X}_d \in \Re^6$ and \(\Dot{\mathcal{X}}_d \in \Re^6\) are the desire pose and desired velocity of end-effector. As it was mentioned, required velocity plays an important role in VDC context, which can be defined based on free-motion or in-contact tasks. In this work, in order to achieve the desired impedance (\ref{des impedance}), the required Cartesian space velocity is defined as,
\begin{equation}\label{DXr}
    \Dot{\mathcal{X}}_r = \Dot{\mathcal{X}}_d + \Gamma\,(\mathcal{X}_d - \mathcal{X}) + \Sigma\,(f_{ed} - \Tilde{f}_e)
\end{equation}
with \(f_{ed}\) being the desired contact force and $\Tilde{f}_e$ being the filtered estimated contact force (\ref{fe_hat}), computed as,
\begin{equation}
    \dot{\Tilde{f}}_e + \mathcal{C}\,\Tilde{f}_e=  \mathcal{C}\,\hat{f}_e,
\end{equation}
 and \(\Gamma \in \Re^{6\times6}\), \(\Sigma \in \Re^{6\times6}\), and \(\mathcal{C} \in \Re^{6\times6}\) being positive-definite matrices. The task space tracking error is \(e = \mathcal{X}_d-\mathcal{X} = [e_p^T,\,e_o^T]^T\), with \(e_p \in \Re^3\) being position error, while \(e_o \in \Re^3\) being orientation error defined based on quaternions as,
\begin{equation}\label{equ39}
     e_o = \eta(q)\epsilon_d-\eta_d\epsilon(q)-S(\epsilon_d)\epsilon(q)
\end{equation}
where \(\eta(q)\) and \(\epsilon(q)\) are unit quaternions computed from the rotation matrix, \(\eta_d\) and \(\epsilon_d\) are desired unit quaternions. As stated in \cite{koivumaki2016stability}, to establish the desired impedance by means of (\ref{DXr}), the following condition must be ensured:
\begin{equation}\label{Gamma}
    \Gamma = K_d\,D_d^{-1},
\end{equation}
\begin{equation}\label{Sigma}
    \Sigma = D_d^{-1}.
\end{equation}
Now, by using Jacobian matrix, the required joint velocity can be achieved:
\begin{equation}\label{dqr}
    \Dot{q}_r = (J)^\dagger\,\Dot{\mathcal{X}}_r.
\end{equation}
The required joint velocity in (\ref{dqr}) conveys the desired impedance to subsystem level, where the control law is designed to establish it, as show in Fig. \ref{Scheme}.

Considering the concept of VPF in VDC, contact only affects the stability of the tool object, which requires attention. Consequently, in the following section, the control design and stability of the tool body will be examined, while the control and stability of the rest of the system is the same as \cite{hejrati2023orchestrated} (Fig. \ref{Scheme} displays the control scheme). This highlights the unique feature of the VDC approach, wherein alterations in one subsystem do not affect the stability of other subsystems.
\begin{figure}[b]
    \centering
    \includegraphics[width=0.3\textwidth]{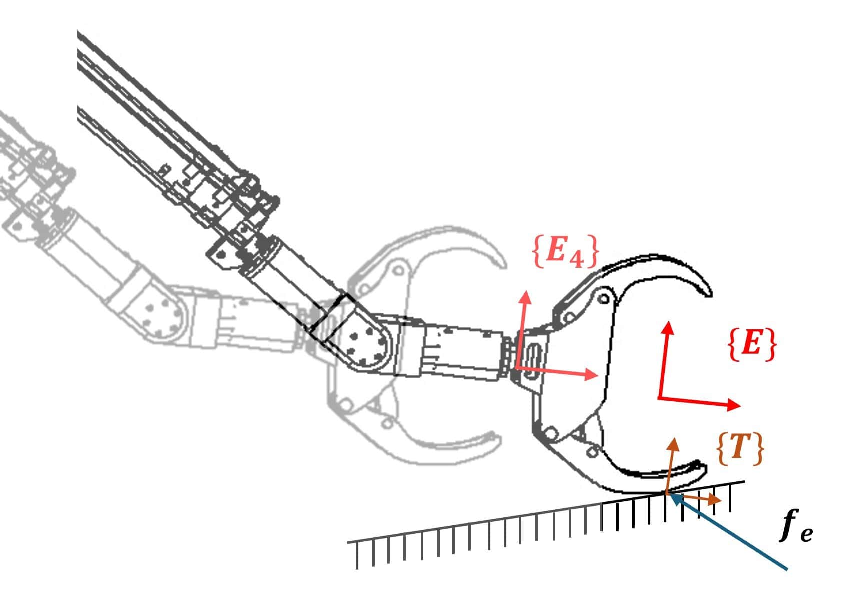}
    \caption{End-effector of HHM approaching an environment}
    \label{Wrist_contact}
\end{figure}

\subsection{Rigid Body Control design}
% Fig. \ref{HIAB} displays the configuration of the HHM along with VDC concept. The HHM in Fig. \ref{fig4} is decomposed into subsystems, as demonstrated in Fig. \ref{fig5}, with actuator mechanisms. On the other hand, Fig. \ref{fig6} depicts the detail of VDC approach with VCP concept. As the new impact resilient control affects the virtual stability of the tool, the analysis is performed on tool body frame $\{E_4\}$ (shown in Fig. \ref{Wrist_contact}).

Considering Fig. \ref{Wrist_contact},  frame $\{T\}$ represents contact point at the end-effector,  frame $\{E\}$ represents the tip of end-effector, and  frame $\{E_4\}$ is the body frame of the tool. Then, the spatial velocity vector of frame $\{T\}$ can be written as,
\begin{equation} \label{TV}
    {^T}V = N_c\, \dot{\mathcal{X}}
\end{equation}
with \(N_c\) mapping the end-effector velocity to the spatial velocity of the tool tip. Then, the spatial velocity of the tool can be derived as,
\begin{equation} \label{E4V}
    {^{E_4}}V = {^{T}}U_{E_4}^T\,{^T}V.
\end{equation}
Accordingly, the spatial force vector of the tool can be written as,
\begin{equation}\label{TF}
    {^T}F = N_c\, f_{e}
\end{equation}
where \(f_e \in \Re^6\) is the contact force between the end-effector and the environment (Fig. \ref{Wrist_contact}). The net spatial force vector $^{E_4}F^*$ of the tool can be written as,
\begin{equation}\label{E4F*}
    M_{E_4} \dfrac{\rm d}{\mathrm{d}t} \left( {^{E_4}V} \right) + C_{E_4} \left( {^{E_4}{\omega}}  \right) {^{E_4}V} + G_{E_4} +\, {^{E_4}}\Delta_R(t) = {^{E_4} F^*}.
\end{equation}
Additionally, the force resultant on the tool body can be written,
\begin{equation}\label{F*E4}
    ^{E_4}F^*  = {^{E_{4}}}F - {^{E_{4}}}U_{T}\, {^T}F,
\end{equation}
By following the same procedure in (\ref{TV})-(\ref{TF}) for required terms, one can obtain,
\begin{equation} \label{TVr}
    {^T}V_r = N_c \Dot{\mathcal{X}}_r,
\end{equation}
\begin{equation} \label{E4Vr}
    {^{E_4}}V_r = {^{T}}U_{E_4}^T\,{^T}V_r,
\end{equation}
\begin{equation} \label{TFr}
    {^T}F_r = N_c\, f_{ed},
\end{equation}
with \(\Dot{\mathcal{X}}_r\) defined in (\ref{DXr}). 
% where the desired contact force corresponding to the flexible environment can be computed as,
% \begin{equation}
% \begin{split}
%     f_{ed} &=\, \sigma \lbrace M_f\, ^T\Dot{V}_{r} + D_f\,^TV_{r} + K_f\,\Tilde{X}_{ef}\rbrace\\
%     & = Y_{e}\,\hat{\theta}_e
% \end{split}
% \end{equation}
% with $\hat{\theta}_e$ being the estimation of the $\theta_e$. On the other side, the desired contact force resulting from rigid environment can be designed as,
% \begin{equation}
%     f_{erd} = \sigma\,A_c^{-1} \left( C_c^{-1}\Ddot{X}_{er} + \Dot{X}_{er}\right).
% \end{equation}
Finally, required net force vector, which acts as the control signal for rigid body, can be expressed in the sense of (\ref{F*r}) as,
\begin{equation}\label{E4F*r}
    ^{E_4}F^*_r = Y_{E_4} \hat{\phi}_{E_4} + K_{E_4}\,(^{E_4}V_r - ^{E_4}V)+\hat{\Delta}_R,
\end{equation}
with \(\hat{\Delta}_R\) being the estimated model uncertainty, as,

\begin{equation}
    \hat{\Delta}_R = \,{^{ {E_4}} }\hat{W}^T\Psi(\chi_{ {E_4}}) +\, {^{ {E_4}}}\hat{\varepsilon}
\end{equation}
where,
\begin{equation}\label{W adapt}
    ^{ {E_4}}\Dot{\hat{W}} = {\Pi}\,\left(\Psi(\chi_{ {E_4}})\,(^{ {E_4}}{ V}_r-\,^{ {E_4}}{ V})^T- ^{ {E_4}}\tau_0\,{^{ {E_4}}\hat{W}}\right)
\end{equation}
\begin{equation}
        {^{ {E_4}}}\Dot{\hat{\varepsilon}} = {\pi} \left (({^{ {E_4}}{ V}_r} - {^{ {E_4}}{ V}})- ^{ {E_4}}\pi_0 {^{ {E_4}}}\hat{\varepsilon}\right),
        \label{eps adapt}
\end{equation}
with \(\hat{W}\) and \(\hat{\varepsilon}\) being the estimation of weights and bias of RBFNNs, \({\pi}\), \({\pi}_0\), and \({\tau}_0\) being positive, \({\Pi}\) being positive definite matrix, \(\Psi(.)\) being Gaussian activation function, and \(\chi_{ {E_4}}\) being the input of the RBFNNs defined in \cite{hejrati2023orchestrated}. As a result, the required spatial force vector can be written as,
\begin{equation}\label{E4F_r}
    {^{E_4}}F_r =\, ^{E_4}F^*_r +  {^{E_{4}}}U_{T}\, {^T}F_r.
\end{equation}

\subsection{Natural Adaptation Law}
For the given rigid body in space, shown in Fig. \ref{Fig bodies}, with an inertial frame \{A\}, there is a unique inertial parameters \(\phi_A \in \Re^{10}\) that satisfy physical consistency conditions. As it is detailed in \cite{hejrati2022decentralized,lee2018natural}, there is a one-to-one linear map \(f:\Re^{10} \rightarrow S(4)\), such that,
\begin{equation*}
    f(\phi_A)= \mathcal{L}_A = \begin{bmatrix}
        0.5tr(I_A).\textbf{1}-I_A & h_A \\
        h^T_A & m_A
        \end{bmatrix}
\end{equation*}
\begin{equation*}
    f^{-1}(\phi_A) = \phi_A(m_A,h_A,tr(\Sigma_A).\textbf{1}-\Sigma_A)
\end{equation*}
where \(m_A\), \(h_A\), and \(I_A\) are the mass, first mass moment, and rotational inertia matrix, respectively. \(\mathcal{L}_A \in S(4)\) is a pseudo-inertia matrix with \(S(4)\) being the space of \(4\times 4\) real-symmetric matrices and \(\Sigma_A = 0.5tr(I_A)-I_A\). The set of physically consistent inertial parameter vectors for a given rigid body can be defined on manifold \(\mathcal{M}\) as,

\begin{equation}\label{M manfold}
\begin{split}
        \mathcal{M} = \{\phi_A \in \Re^{10}: f(\phi_A)\succ0\} \subset \Re^{10} \\
        = \{\mathcal{L}_A \in S(4): \mathcal{L}_A\succ0\}=\mathcal{P}(4)
\end{split}
\end{equation}
where \(\mathcal{P}(4)\) is the space of all real-symmetric, positive-definite matrices. According to (\ref{M manfold}), for a given inertial parameter vector \(\phi_A\), if \(\mathcal{L}_A \in \mathcal{P}(4)\), then the physical consistency condition is satisfied.

For a given \(\mathcal{L}_{A}\) with its estimation \(\mathcal{\hat{L}}_{A}\), the Lyapunov function can be defined in the form of Bregman divergence with the log-det function as,
\begin{equation} \label{Df}
    \mathcal{D}_F(\mathcal{L}_{A}\rVert \hat{\mathcal{L}}_{A}) = log\frac{|\hat{\mathcal{L}}_{A}|}{|\mathcal{L}_{A}|}+tr(\hat{\mathcal{L}}_{A}^{-1}\mathcal{L}_{A})-4.
\end{equation}
with,
\begin{equation*}
    \Dot{\mathcal{D}}_F(\mathcal{L}_{A}\rVert \hat{\mathcal{L}}_{A}) = tr([\hat{\mathcal{L}}_{A}^{-1}\Dot{\hat{\mathcal{L}}}_{A}\,\hat{\mathcal{L}}_{A}^{-1}]\,\Tilde{\mathcal{L}}_{A})
\end{equation*}
being the time-derivative of (\ref{Df}), where \(\Tilde{\mathcal{L}}_{A} = \hat{\mathcal{L}}_{A} - \mathcal{L}_{A}\). Thus, the natural adaptation law (NAL) can be derived as,
\begin{equation}
        \Dot{\hat{\mathcal{L}}}_A = \frac{1}{\gamma}\, \hat{\mathcal{L}}_A\,\left(\mathcal{S}_A-\gamma_0\,\hat{\mathcal{L}}_A\right)\, \hat{\mathcal{L}}_A
        \label{L adapt}
\end{equation}
with \(\gamma >0\) being the adaptation gain for all the rigid bodies of the system, and \(\gamma_0>0\) being a small positive gain. Additionally, \(\mathcal{S}_A\) is a unique symmetric matrix defined in \cite{hejrati2023physical}.

\begin{rmk}
    Exploiting NAL defined in (\ref{L adapt}) along with Property \ref{property 1}, the physical consistency of all inertial parameters of the system, and, consequently, the positive-definiteness of the estimated spatial inertia matrix will be guaranteed.
\end{rmk}

\subsection{System Stability}
Due to environmental contact, the second term in (\ref{F*E4}) is non-zero, differing from the free motion condition in \cite{hejrati2023orchestrated}. Consequently, only the virtual stability of the tool needs to be ensured under these contact conditions, while the stability of the rest of the system is the same as in \cite{hejrati2023orchestrated}. 
% This exemplifies the distinctive advantage of VDC approach, wherein changes in one subsystem do not impact the stability of others \cite{zhu2010virtual}.
% Due to the contact with environment, the second term in (\ref{E4F_r}) is not zeros in contrast to the condition in \cite{hejrati2023orchestrated}. Therefore, only the virtual stability of the tool needs to be ensured in presence of the contact, and the stability of the rest of the system remains the same. This demonstrates the perfect property of the VDC in which change in a subsystem, does not affect other subsystems \cite{zhu2010virtual}.

\begin{thm}
    Consider the tool body with governing dynamics of (\ref{E4F*}), control action of (\ref{E4F*r}), and adaptation law of (\ref{L adapt}) by replacing \{A\} with $\{E_4\}$. By defining the accompanying function in the sense of Definition \ref{nu def} as,
     \begin{equation}
    \begin{split}
	{\nu}_1 &=  \dfrac{1}{2} \, \left( {^{ {E_4}}{ V}_r} - {^{ {E_4}}{ V}}  \right)^T \, {\rm M_{\rm {E_4}}} \, \left( {^{ {E_4}}{ V}_r} - {^{ {E_4}}{ V}}  \right)\\
    &+\, \gamma \mathcal{D}_F(\mathcal{L}_{{E_4}}\rVert \hat{\mathcal{L}}_{{E_4}}) + \frac{1}{2}tr({^{ {E_4}}\Tilde{W}^T}\,{^{ {E_4}}\Gamma^{-1}}{^{ {E_4}}\Tilde{W}}) \\
        &+ \frac{1}{2\,{^{ {E_4}}}\pi} {^{ {E_4}}\Tilde{\varepsilon}^T}\,{^{ {E_4}}\Tilde{\varepsilon}},
    \end{split}
	\label{eqn: v function for RB}
    \end{equation}
    where \(\Tilde{(.)} = (.) - \hat{(.)}\), and its time derivative,
    \begin{equation}\label{dnu2}
    \Dot{{\nu}}_1 \leq -\alpha_1 {\nu}_1 + \alpha_{10}+p_{E_4}-p_T
    \end{equation} 
    the virtual SGUUB of the tool body subsystem can be ensured. %with $\alpha_1$ and $\alpha_{10}$ being positive.
\label{thm: rigid body}
\end{thm}
\begin{pf}
    Taking the time derivative of (\ref{eqn: v function for RB}) and using (\ref{E4F_r}), (\ref{E4F*}), (\ref{E4F*r}), (\ref{W adapt}), (\ref{eps adapt}), (\ref{L adapt}), and following the same procedure in Appendix A of \cite{hejrati2023orchestrated}, one can obtain (\ref{dnu2}).
\end{pf}

As it can be seen from (\ref{dnu2}) and Fig. \ref{fig6}, the \(p_{E_4}\) is the VPF from the preceding subsystem, while \(p_T\) is VPF resulting from end-effector interaction with environment, which is addressed in the following lemma.
\begin{lem}
    Defining the impedance control command in the sense of (\ref{DXr}) with (\ref{Gamma}) and (\ref{Sigma}) results in,
    \begin{equation*}
        p_T = 0.
    \end{equation*}
\end{lem}
\begin{pf}
    Appendix B.
\end{pf}
% \begin{rmk}
%     The voltage control term in (\ref{ufr}) is designed to ensure the joint tracking accuracy and to generate the required spatial force computed in (\ref{E4F*r}). This controller is applied to all six actuators of the system as they have similar.
% \end{rmk}
\begin{thm}
    Consider the HHM in Fig. \ref{fig4} which is decomposed into subsystem in Fig. \ref{fig6}. The entire system is SGUUB under impedance control law (\ref{DXr}) with all the adaptation laws. 
    \label{General thm}
\end{thm}
\begin{pf}
    Appendix C.
\end{pf}
% As a result, the impact resilient control law along with proposed force estimator has elaborated on in this section and stability analysis are accomplished. In the following section, the experimental results are provided. %The Algorithm. \ref{alg:alg1} illustrates the procedure of implementation of the approach on the hardware.

\section{Result}
This section presents the experimental results of the designed impact-resilient controller in both free motion and contact operations. Figure \ref{Scheme} illustrates the control scheme employed in the experimental implementation.
\begin{figure}[t]
    \centering
    \includegraphics[width=0.4\textwidth]{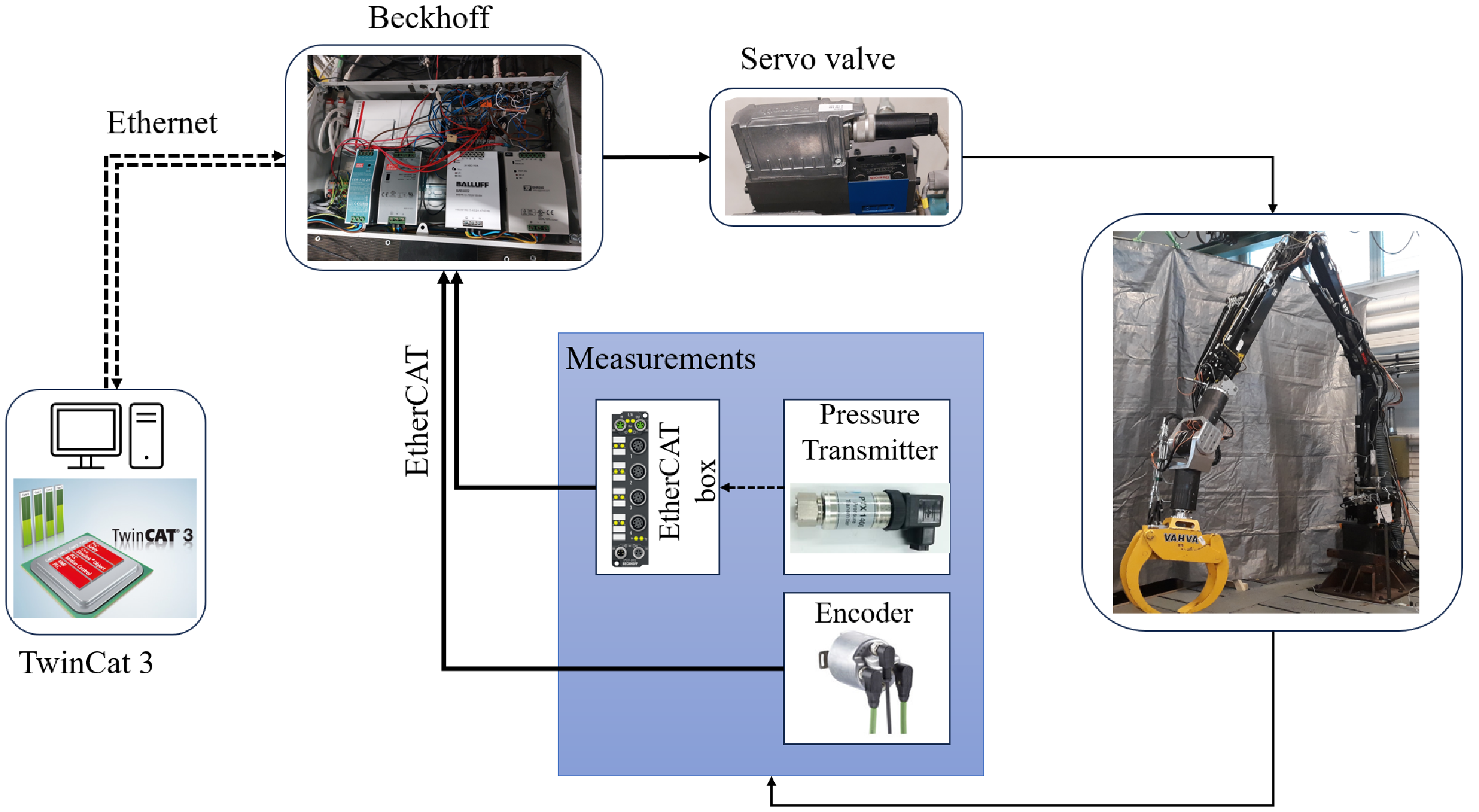}
    \caption{Experimental setup}
    \label{set up}
\end{figure}

\subsection{Experimental Setup}
Fig. \ref{set up} demonstrates the setup for performance evaluation of the controller. Druck PTX1400 and Unik 5000 pressure transmitters (range 25 MPa) and Sick afS60 (18-bit) absolute encoders are utilized in the experiments. Additionally, EP3174-0002 EtherCAT box has been utilized to convert the analog pressure data to digital data, which can be read by Beckhoff platform. The Bosch Rexroth NG6 size servo solenoid valve with 12 l/min for base actuator, 100 l/min for second and third actuators, and 40 l/min for the Eckart wrist, all at $\Delta P = 3.5 MPa$ per notch, are utilized to control the flow. A couple of wooden pallets are utilized to represent the unknown environment.  The entire controller is designed in the host PC with Intel Core i7-6700 CPU 3.40 GHz, and uploaded to the Beckhoff platform with Intel Core i7 2610UE 1.5 GHz for real-time implementation. The human-machine-interface has been designed in TwinCat 3 to enable the communications with 1 ms sample time. A smooth fifth-order trajectory generator is utilized to produce a smooth trajectory between the set points for a given execution time, $t_f$. The smaller the $t_f$, the faster the trajectory.

\begin{figure}[t]
    \centering
    \includegraphics[width=0.5\textwidth]{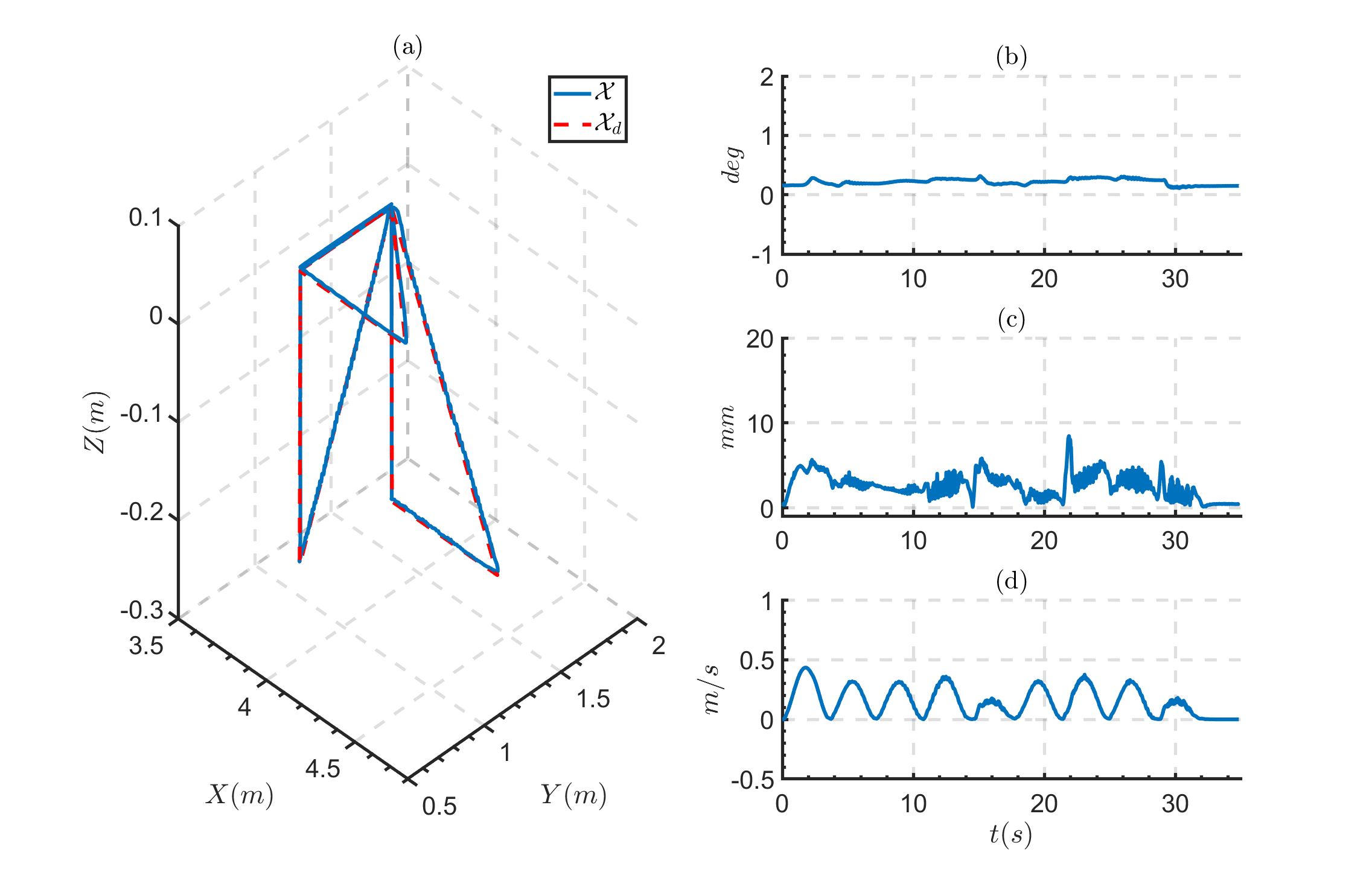}
    \caption{Path tracking with \(t_f = 3.5s\). a) Desired pose tracking, b) RMS of orientation error, c) RMS of position error, d) norm of end-effector velocity.}
    \label{all_3_5}
\end{figure}
\begin{figure}[b]
    \centering
    \includegraphics[width=0.5\textwidth]{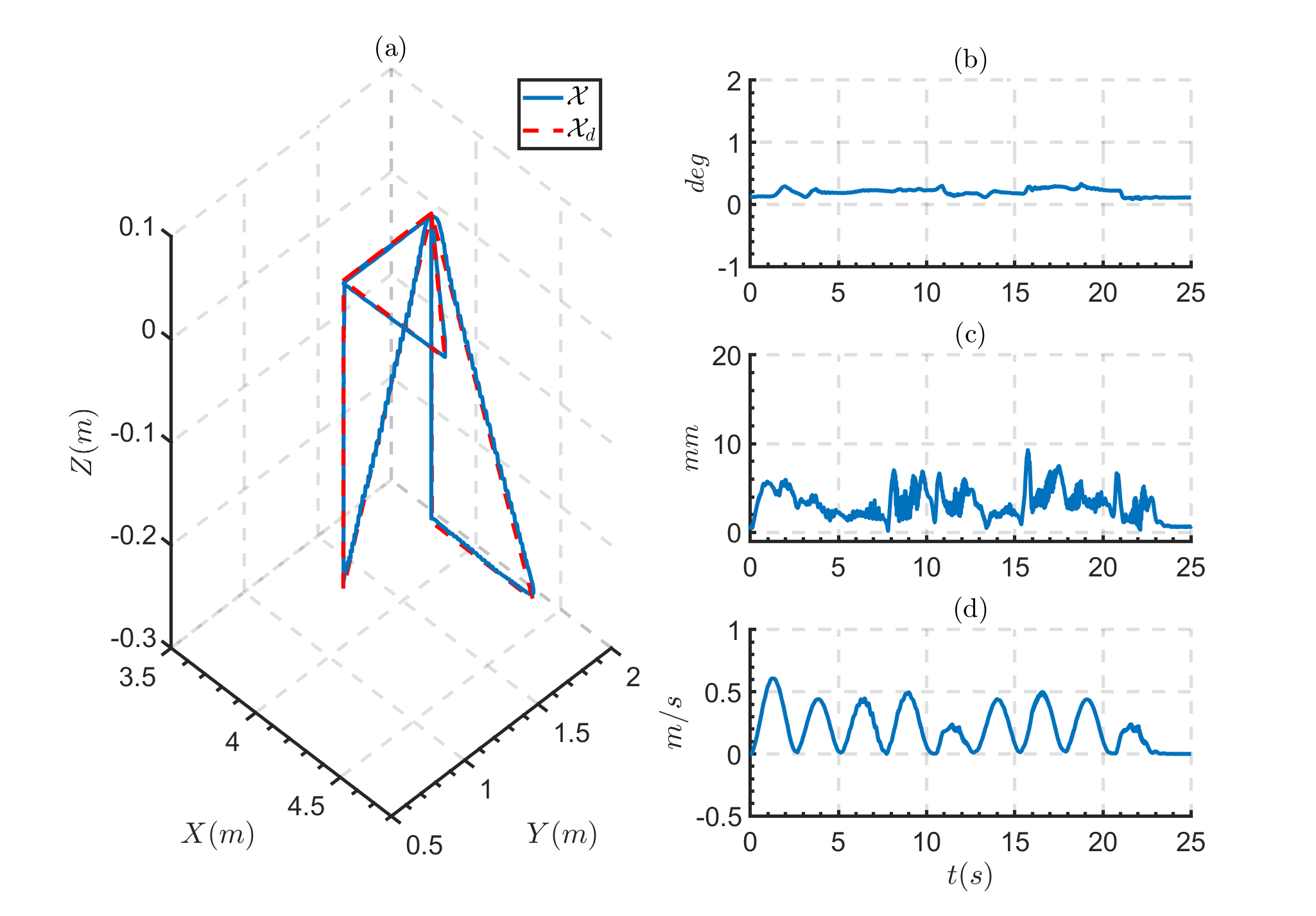}
    \caption{Path tracking with \(t_f = 2.5s\). a) Desired pose tracking, b) RMS of orientation error, c) RMS of position error, d) norm of end-effector velocity.}
    \label{all_2_5}
\end{figure}

\subsection{Free Motion Performance}
In order to evaluate the performance of the controller in free motion, a 3D triangular path with different velocities (different \(t_f\)) has been designed. The control gains have been selected as follows to get the best tracking result: \(K_d = [1,0.9,0.7,1.2,1.2,1.2]\cdot10^6\),\,\(D_d = [1,1,1,1,1,1]\cdot10^5\), \(K_{E4} = 50\cdot I\), \(\gamma = 500\), \(\Pi = 300\), \(\pi = 10\), and low level control gains as in \cite{hejrati2023orchestrated}. 
\newpage
Fig. \ref{all_3_5}(a) demonstrates the path tracking performance with the desired values in Cartesian space for \(t_f = 3.5s\). Fig. \ref{all_3_5}(b) shows the root-mean-square (RMS) of orientation error, Fig. \ref{all_3_5}(c) depicts the RMS of position error, and Fig. \ref{all_3_5}(d) displays the norm of end-effector velocity. As can be seen, the total RMS of less than a 0.5 degree for orientation and less than 1 cm for position has been achieved. Moreover, Fig. \ref{all_2_5} displays the same path tracking performance for \(t_f = 2.5s\), the faster trajectory. It can be seen from Fig. \ref{all_2_5}(d) that although the velocity of the end-effector has been increased, the subcentimeter error has been preserved for position.

\begin{figure}[t]
    \centering
    \includegraphics[width=0.4\textwidth]{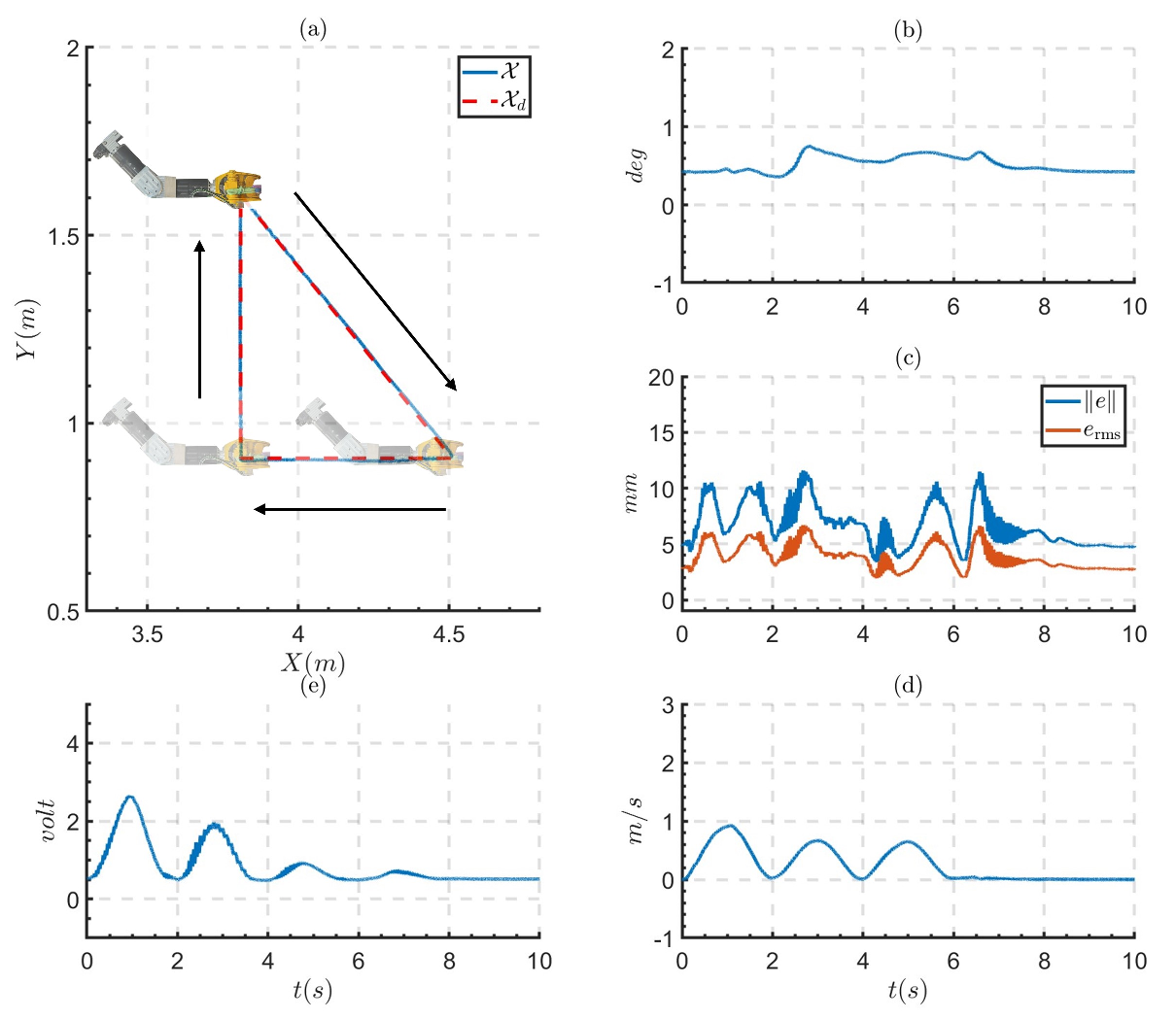}
    \caption{Path tracking with \(t_f = 1.5s\). a) desired pose tracking, b) RMS of orientation error, c) Norm and RMS of position error with blue and red line, respectively, d) Norm of end-effector velocity, e) Norm of voltage command.}
    \label{All_in}
\end{figure}

\begin{table}[b]
\centering
\caption{Performance index evaluation of Cartesian space}
\label{table rho}
\setlength{\tabcolsep}{10pt}
\begin{threeparttable}
\begin{tabular}{ P{70pt} P{20pt} P{10pt} P{20pt} }
\hline
Study&  $\rho$ &  DoF$^\star$ &  $\rho$/DoF \\ [0.5ex]
\hline \hline
This study &
\textbf{0.0124}&
\textbf{6}&
\textbf{0.0021}\\
Koivumaki 2015 \cite{koivumaki2015high} &
0.005 &
2$^\diamond$ &
0.0025\\
Koivumaki 2019 \cite{koivumaki2019energy} &
0.0057 &
2$^\diamond$ &
0.0029\\
Kalmari 2015 & 
0.12$^\top$&
3$^\ddagger$&
0.04\\
Zhu 2005 & 
0.015$^\top$&
3$^\ddagger$&
0.005\\
Tsukamoto 2002 & 
0.126$^\top$&
6&
0.021\\
\hline
\end{tabular}
\begin{tablenotes}
\scriptsize
\item[$\star$] DoF in Cartesian space with 3 DoF for position and 3 DoF for orientation.
\item[$^\diamond$] Only x- and y- directions were considered.
\item[$^\top$] Reported in \cite{mattila2017survey}.
\item[$^\ddagger$] Only position of end-effector was considered.
\end{tablenotes}
\end{threeparttable}
\label{tab}
\end{table}

One of the well-known indexes to evaluate the performance of the controller in free motion tasks is \(\rho\) value \cite{mattila2017survey},
\begin{equation}
    \rho = \frac{|e|_{max}}{|\Dot{\mathcal{X}}|_{max}}
\end{equation}
which takes the velocity of the manipulator at the time of tracking into account. The \(\rho\) value indicates the importance of low tracking error in higher velocities, which is utilized to compare the result of this study with previous studies. Fig. \ref{All_in}(a) illustrates the result of fast trajectory tracking (\(t_f = 1.5s\)) of triangular path in x-y plane while orientation and z-direction position are active and in set-point control mode. Fig. \ref{All_in}(b) depicts the RMS error of orientation regulation (which is still less than a degree), Fig. \ref{All_in}(c) shows the RMS (red line) and norm (blue line) of the position error, the norm of the end-effector velocity is exhibited in Fig. \ref{All_in}(d), and the norm of the voltage commands to all the valves is demonstrated in Fig. \ref{All_in}(e). The maximum norm of the position error is 11.5mm at the maximum velocity of the 0.93 m/s, resulting in \(\rho = 0.0124s\). Note that the maximum value of RMS error is almost 5 mm. Table \ref{table rho} provides the comparison of the results with previous studies based on the number of active Cartesian space DoFs. As can be seen, the achieved result of the present paper outperforms other approaches. Establishing such a result for a real-world HHM (with weight of 650 kg at 4 meters reach) demonstrates the excellent performance of the proposed method and its universality and reliability in real-world applications. 
% In our experiments, in order to achieve high velocity for end-effector, we limited the tracking in x-y plane (which is derived by high-flow-rate valves), while all other joint are active and regulating orientation and z-direction position.
% Additionally, the \(\rho\) value achieved in this study is almost the same as our previous work \cite{hejrati2023orchestrated} which shows the repeatability of the achieved results.

\begin{figure}[t]
    \centering
    \includegraphics[width=0.35\textwidth]{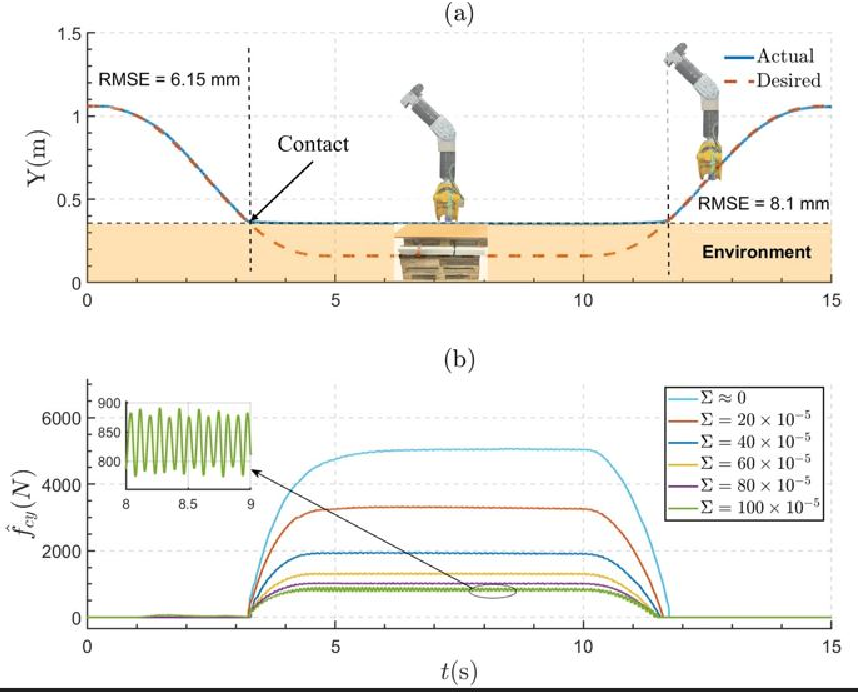}
    \caption{Estimated impact force with various \(\Sigma\). a) path tracking in y direction with contact, and, b) estimated forces.}
    \label{lam_f}
\end{figure} 

\subsection{In-contact Performance}
In this section, the performance of the controller in contact-rich operations is experimentally investigated. The control gains are the same as previous section with \(K_d = diag([1,0.0075,0.7,1.2,1.2,1.2])\cdot10^6\). First of all,to evaluate the accuracy of the proposed force estimator, the simulation result is used. The Simscape toolbox of SIMULINK is utilize to model the contact of the HHM with an environment. The achieved RMS of error between estimated force and reported force from Simscape is 41.8 N with steady-state error of 18 N, showing the acceptable accuracy of the force estimator. In the following, the results of extensive experiments of implementing the designed method on a real-world HHM are provided.

As the \(\Sigma\) in (\ref{DXr}) reflects the effect of contact, we first experimentally evaluate its effect. Fig. \ref{lam_f} depicts the result of contact in y direction with wrist down configuration, which is the configuration to pick up an object (relevant application in real-world cases). In this case, the frame \{T\} and \{E\} of Fig. \ref{Wrist_contact} overlap. As can be seen from Fig. \ref{lam_f}, higher values of \(\Sigma\) will result in instability of the contact, while its absence will impose a harsh damage to both manipulator and the environment. According to the Fig. \ref{lam_f}, the \(\Sigma = [1,60,1,1,1,1]^T \cdot 10^{-5}\) is suitable for the following experiment (corresponding to \(D_d = diag([1,0.0167,1,1,1,1])\cdot 10^{5}\)), accounting for safety and stability. It is worth noting the subcentimeter tracking accuracy of the proposed controller before and after contact. Additionally, as illustrated in Fig. \ref{lam_f}(b), an appropriate selection of \(\Sigma\) can significantly reduce the contact force from 5 kN  (\(\Sigma = 0\)) to 1 kN (\(\Sigma = 100e-5\)), corresponding to an 80 \% reduction in the impact of the contact. This is a substantial achievement, especially for HHMs in heavy-duty operations.

\begin{figure}
    \centering
    \includegraphics[width=0.5\textwidth]{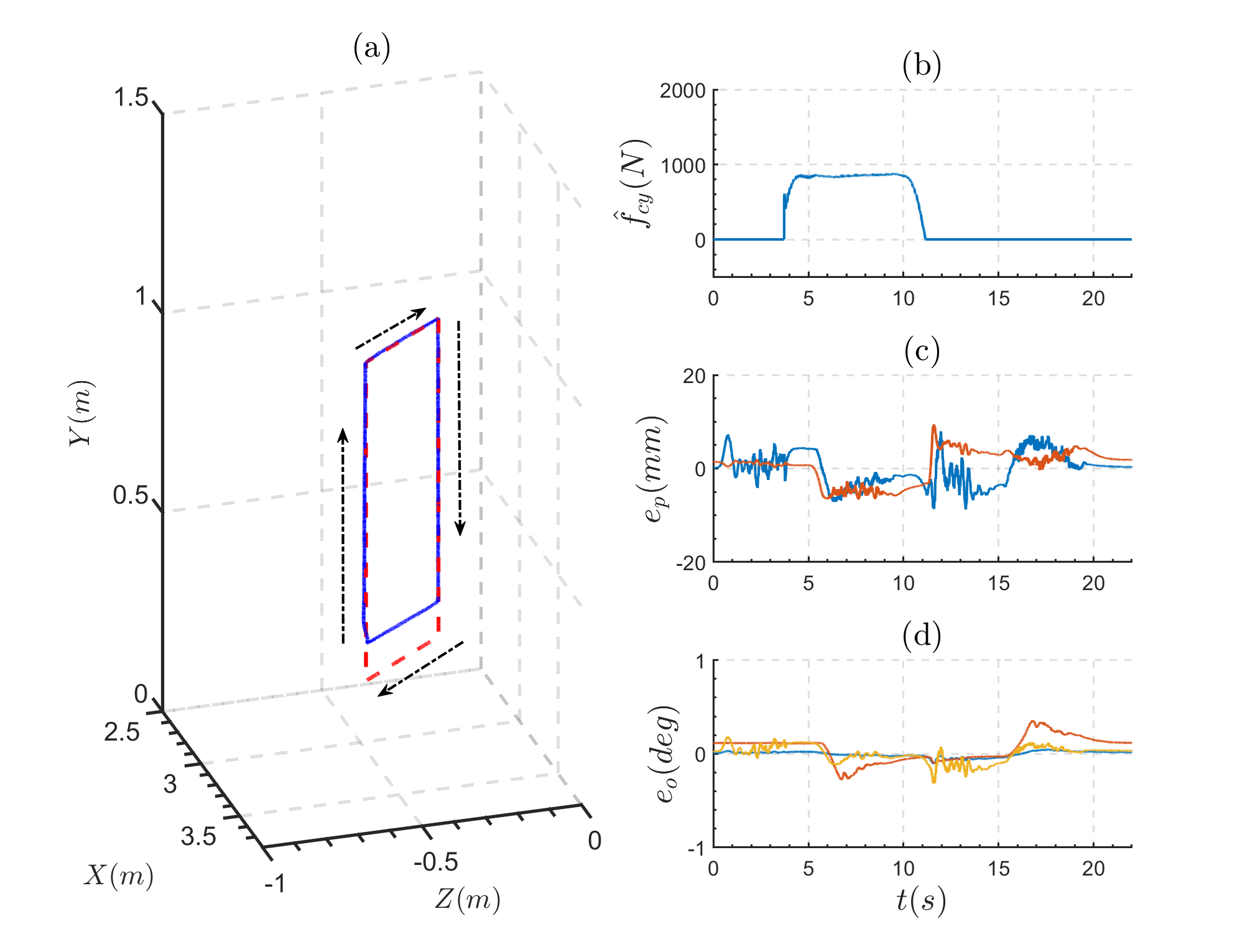}
    \caption{a) Path tracking in Cartesian space with contact, b) Impact force, c) Position error in non-contact directions, and, d) Orientation error.}
    \label{cont_5s}
\end{figure}

\begin{figure}
    \centering
    \includegraphics[width=0.5\textwidth]{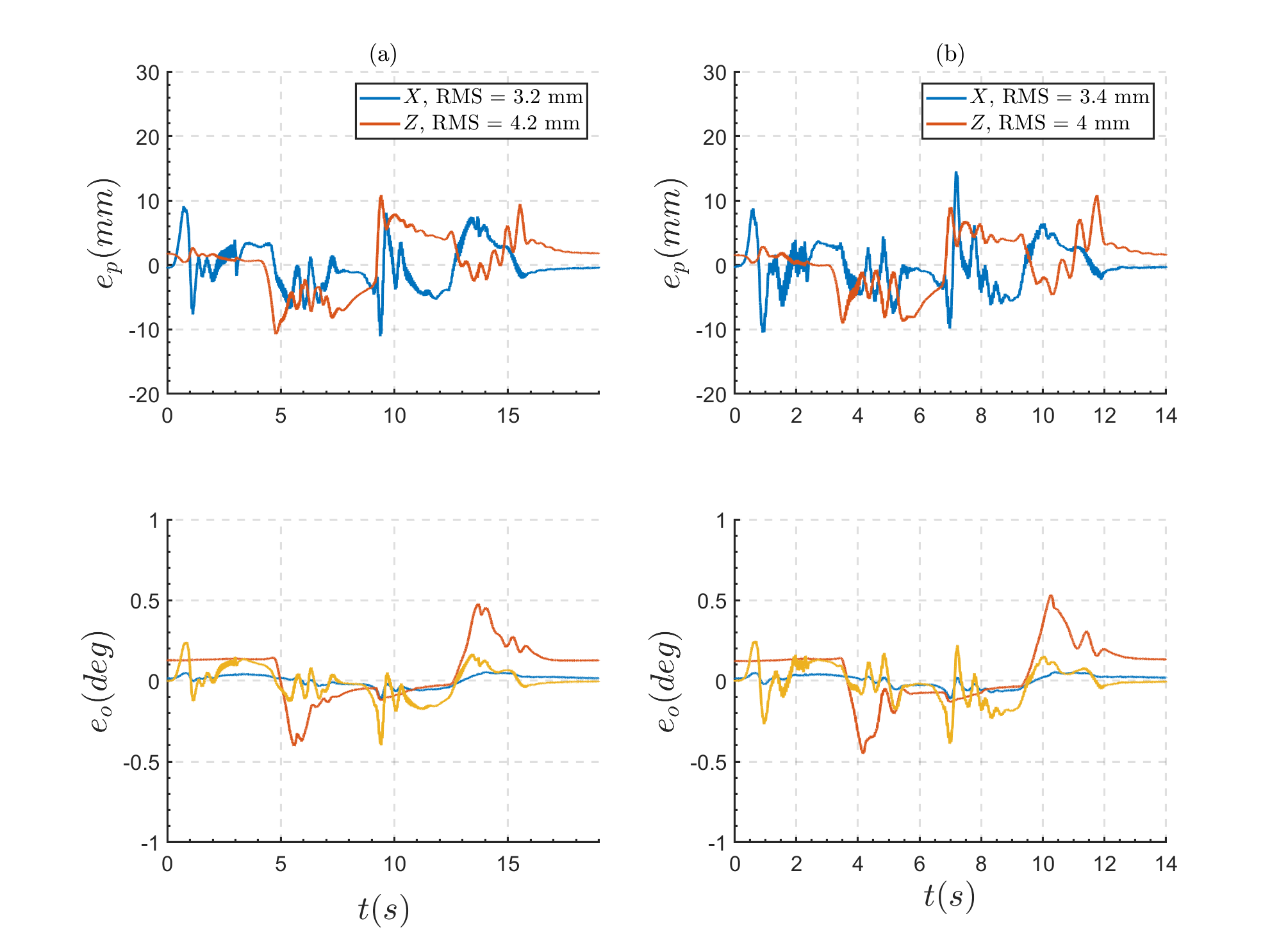}
    \caption{Position and orientation tracking during contact. a) Position and orientation errors for \(t_f = 4s\), and, b) Position and orientation errors for \(t_f = 3s\).}
    \label{tf 43s}
\end{figure}

Fig. \ref{cont_5s}(a) illustrates the pose tracking of the end-effector with an unexpected contact with unknown environment. As can be seen, the tracking error in Fig. \ref{cont_5s}(c) in non-contact directions (x and z) are less than a centimeter for \(t_f = 5s\) and orientation error in Fig. \ref{cont_5s}(d) is less than a 0.5 degree. Fig. \ref{cont_5s}(b) also shows the estimated contact force in the contact direction (\(\hat{f}_{cy}\)). In order to display the robustness and stability of the controller, the same experiment with \(t_f = 4s\) and \(t_f = 3s\), which are faster trajectories, are performed. As can be seen from Fig. \ref{tf 43s}, the orientation and position error in non-contact directions are less than a 0.5 degree and 1 centimeter, respectively, despite of following a fast trajectory in presence of contact. All the provided results illustrated the perfect performance of the controller in preserving safe contact while accomplishing an excellent tracking in non-contact direction.

\begin{figure}
    \centering
    \includegraphics[width=0.4\textwidth]{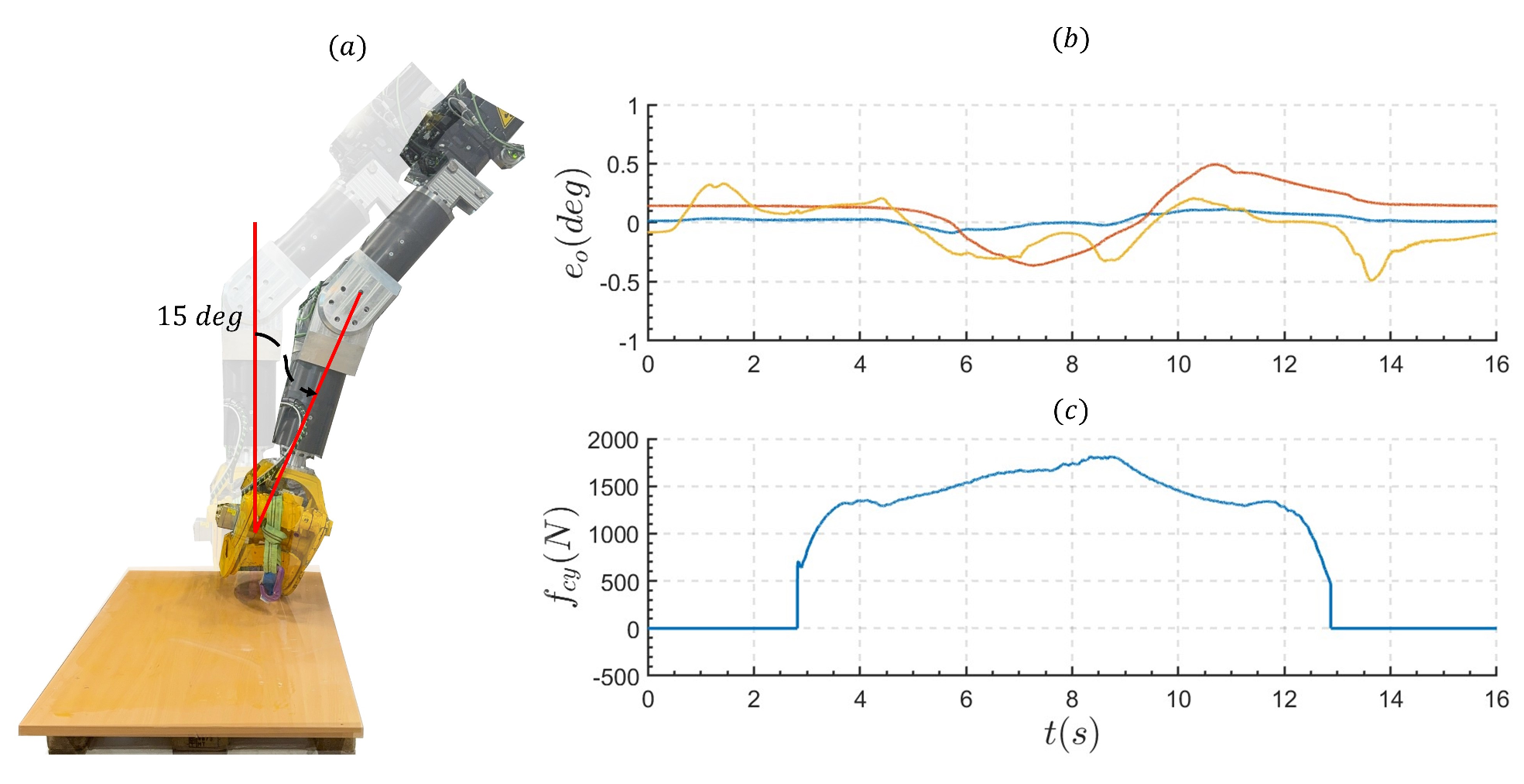}
    \caption{a) Wrist maneuvering during the contact with \(t_f = 4s\), b) Orientation error, and c) Contact force.}
    \label{xy}
\end{figure}

\begin{figure}
    \centering
    \includegraphics[width=0.4\textwidth]{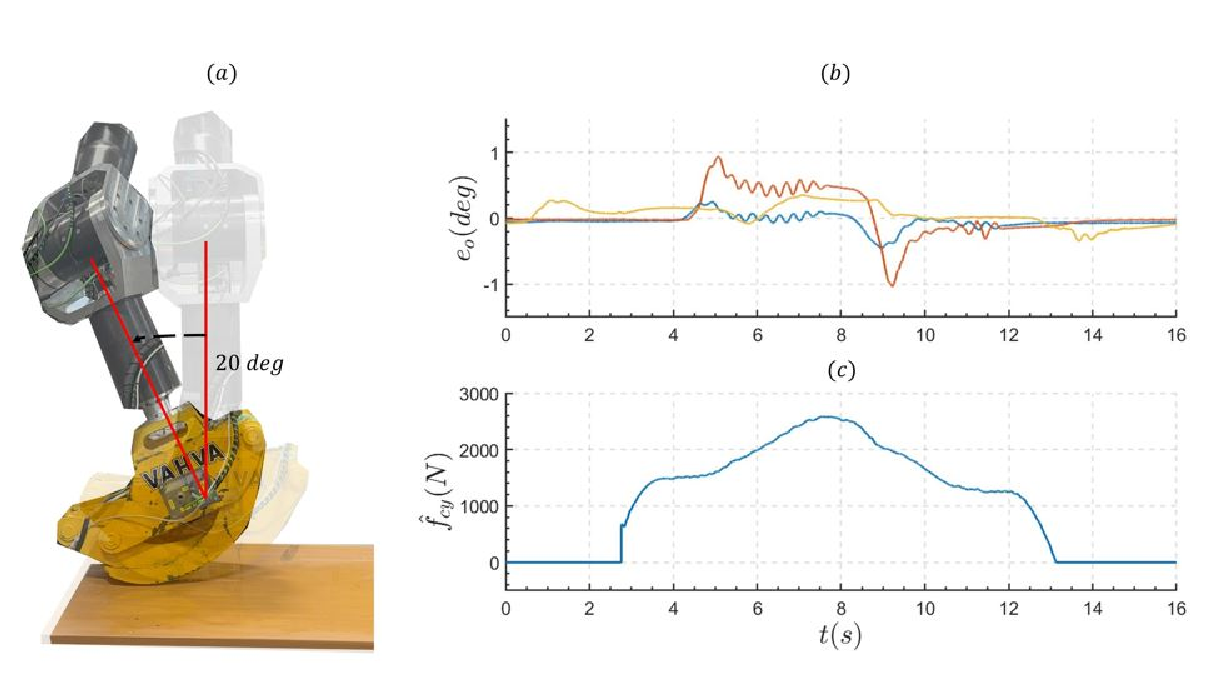}
    \caption{a) Wrist maneuvering during the contact with \(t_f = 4s\), b) Orientation error, and c) Contact force.}
    \label{xz}
\end{figure}

% Fig. \ref{lam_f} illustrates the path tracking in the y direction with unexpected contact with unknown environment. As it can be seen, the contact does not result in instability and the designed controller perfectly tackled the issue, and achieved an accurate tracking for orientation and position, as can be seen from Fig. \ref{ep6s} and Fig. \ref{eo6s}. The orientation error is almost under 1 degree with position errors in non-contact directions under 5mm. Additionally, Fig. \ref{F_hat} depicts the estimated contact force using the proposed algorithm. The estimated contact force eliminated the need for force sensors to achieve safe contact.

In many contact-rich scenarios, dexterous operations may be required while maintaining contact with the environment. Successfully performing such tasks necessitates establishing safe contact and achieving precise control accuracy. Figures \ref{xy} and \ref{xz} examine the wrist maneuvering of the HHM during contact. As shown, excellent orientation accuracy is achieved across different maneuvering scenarios, emphasizing good performance of the controller. 

\begin{figure}[h!]
    \centering
    \includegraphics[width=0.45\textwidth]{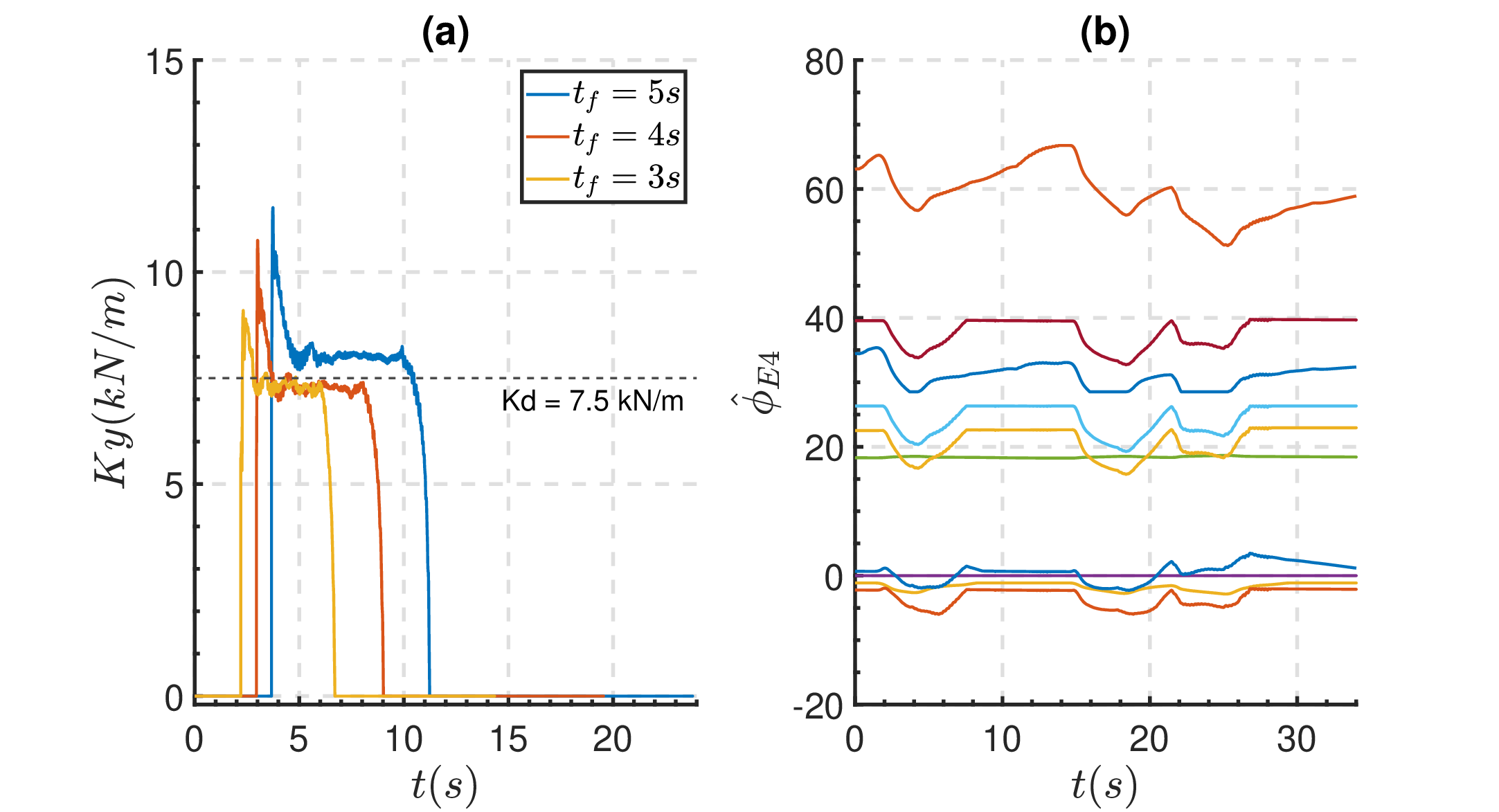}
    \caption{a) Rendered stiffness in contact direction, b) time history of \(\hat{\phi}_{E_4}\)}
    \label{k}
\end{figure}

Additionally, Fig. \ref{k}(a) presents the time history of the stiffness rendered by the manipulator during contact at different velocities. It can be observed that the manipulator's behavior during contact closely matches the desired stiffness  (\(K_y \cong K_{dy}\)) with  accuracies of \(6\%\), \(2\%\), and \(2\%\) for \(t_f = 5s\) and \(4s\), and \(3s\), respectively, indicating successful rendering of the desired impedance as defined in (\ref{des impedance}). Moreover, the adaptation signals for the unknown rigid body parameters are shown in Fig. \ref{k}(b), demonstrating the boundedness of the estimated parameters.

\section{Conclusion}
In this work, an impact-resilient control method is proposed to achieve high accuracy in path tracking during free motion and to reduce damage from unexpected impacts for a generic 6-DoF HHM. Additionally, a novel representation of the GMO is introduced in Plücker coordinates for the first time, aligning with the dynamics of the proposed scheme. The stability of the system under the new control command is demonstrated using the VPF and the virtual stability concepts of VDC. Extensive experiments have been conducted to evaluate the performance of the proposed scheme. The results show that the designed impact-resilient approach achieves subcentimeter tracking accuracy and reduces the impact of contact by 80\%.

\section{Appendix}
\subsection{ Derivative of inertia matrix}
Spatial inertia matrix in (\ref{M_A}) is a symmetric dyadic tensor that maps \(\mathcal{M}^6\) to \(\mathcal{F}^6\). Consequently, it can be expressed as the sum of six symmetric dyads as \cite{featherstone2014rigid},
\begin{equation}
    M_A = \sum_{i=1}^6 {\mathfrak{m}_i\,\mathfrak{m}_i \bullet} \quad \mathfrak{m}_i \in \mathcal{F}^6.
\end{equation}
with \(\mathfrak{m}_i \bullet = \mathfrak{m}_i ^T\). It must be mention that \(M_A\) is fixed to the body and only varies with the body motion, allowing to represent it as a sum of dyads in which each of \(\mathfrak{m}_i\) are all fixed to the corresponding body. Consequently, we can define \(\times\) operator for 6D vectors as \(\Dot{\mathfrak{m}}_i =\,  ^A\mathcal{V}\times^*\,\mathfrak{m}\) with \(^A\mathcal{V}\times^* = -(^A\mathcal{V}\times)^T\), and
% \begin{equation}
%     \Dot{\mathfrak{m}}_i =\,  ^A\mathcal{V}\times^*\,\mathfrak{m}
% \end{equation}
% with \(^A\mathcal{V}\times^* = -(^A\mathcal{V}\times)^T\), and ,
\begin{equation}\label{V*}
^A\mathcal{V}\times = \begin{bmatrix}
^A\omega\times & ^Av\times \\
0& ^A\omega\times
\end{bmatrix},
\quad
\alpha\times = \begin{bmatrix}
0 & -\alpha_z & \alpha_y \\
\alpha_z & 0 & -\alpha_x\\
-\alpha_y & \alpha_x & 0
\end{bmatrix},
\end{equation}
where \(\alpha = [\alpha_x,\alpha_y,\alpha_z] \in R^3\) is any vector. Then, the derivative of \(M_A\) can be computed as follow:
\begin{equation}
\begin{split}
    \frac{d}{dt}M_A &= \sum_{i=1}^6 {\Dot{\mathfrak{m}}_i\,\mathfrak{m}_i \bullet +\mathfrak{m}_i\, \Dot{\mathfrak{m}}_i\bullet}\\
    &=  \sum_{i=1}^6{(^AV\times^*\mathfrak{m}_i)\mathfrak{m}_i \bullet} +  \sum_{i=1}^6{\mathfrak{m}_i\,(^AV\times^*\,\mathfrak{m}_i)} \bullet\\
    &=\, ^AV\times^*\,\sum_{i=1}^6 {\mathfrak{m}_i\,\mathfrak{m}_i \bullet} - (\sum_{i=1}^6 {\mathfrak{m}_i\,\mathfrak{m}_i \bullet})\,^AV\times\\
    & =\, -(^AV\times)^T\,M_A - M_A\,^AV\times.
\end{split}
\end{equation}

\subsection{Lemma 1}
By recalling Definition (\ref{VPF}), (\ref{TV}), (\ref{TF}), (\ref{TVr}), (\ref{TFr}), (\ref{DXr}), (\ref{des impedance}), and \(N_c^{T}\,N_c = I_6\), one can obtain,
\begin{equation} \label{p_T}
\begin{split}
    p_T &= (^TV_r -\, ^TV)^T\,(^TF_r -\, ^TF) \\
    % & =  (\Dot{\mathcal{X}}_r - \Dot{\mathcal{X}})^T\,N_c^{T}\,N_c\,(f_{ed} - f_e)\\
    & =  ((\Dot{\mathcal{X}}_d-\Dot{\mathcal{X}}) + \Gamma\,(\mathcal{X}_d - \mathcal{X}) + \Sigma\,(f_{ed} - \Tilde{f}_e))^T\,(f_{ed} - f_e)\\
    & = -(\Dot{\mathcal{X}}_d-\Dot{\mathcal{X}})^T\,(D_d\Sigma\,D_d-D_d)(\Dot{\mathcal{X}}_d-\Dot{\mathcal{X}})\\
    &+ (\mathcal{X}_d-\mathcal{X})^T\,(K_d\,\Sigma\,K_d -\Gamma\,K_d)(\mathcal{X}_d-\mathcal{X})\\
    &+(\mathcal{X}_d-\mathcal{X})^T(2D_d\Sigma\,K_d-\Gamma\,D_d-K_d)(\Dot{\mathcal{X}}_r - \Dot{\mathcal{X}}).
\end{split}
\end{equation}
It can be seen from (\ref{p_T}) that selecting the \(\Gamma\) and \(\Sigma\) as is in (\ref{Gamma}) and (\ref{Sigma}), respectively, yields in \(p_T = 0\).
\subsection{Proof of Theorem \ref{General thm}}
Consider the accompanying function for the system with tool body excluded as \(\nu_R\) with its time derivative as,
\begin{equation} \label{dnuR}
    \dot{\nu}_R \leq -\Bar{\alpha}_1\nu_R+\Bar{\alpha}_0-p_{E_4}
\end{equation}
with \(p_{E_4}\) being VPF at the driven cutting point (Fig. \ref{fig6}). Let \( \nu_T = \nu_1 + \nu_R\) be the accompanying function of the entire system. Then, using (\ref{dnu2}) and (\ref{dnuR}) along with Lemma 1 with following the same procedure in \cite{hejrati2023orchestrated}, one can achieve,
\begin{equation}
    \dot{\nu}_T \leq -\mu\,\nu_T+\mu_0
\end{equation}
which ensures the SGUUB for entire system in the sense of Theorem \ref{thm1}.

\bibliographystyle{IEEEtran}
\bibliography{mybib}

% \newpage

% \section{Biography Section}
 
% \vspace{11pt}

% \bf{If you include a photo:}\vspace{-33pt}
% \begin{IEEEbiography}[{\includegraphics[width=1in,height=1.25in,clip,keepaspectratio]{fig1}}]{Michael Shell}
% Use $\backslash${\tt{begin\{IEEEbiography\}}} and then for the 1st argument use $\backslash${\tt{includegraphics}} to declare and link the author photo.
% Use the author name as the 3rd argument followed by the biography text.
% \end{IEEEbiography}

% \vspace{11pt}

% \bf{If you will not include a photo:}\vspace{-33pt}
% \begin{IEEEbiographynophoto}{John Doe}
% Use $\backslash${\tt{begin\{IEEEbiographynophoto\}}} and the author name as the argument followed by the biography text.
% \end{IEEEbiographynophoto}

\begin{IEEEbiography}[{\includegraphics[width=1in,height=1.25in,clip,keepaspectratio]{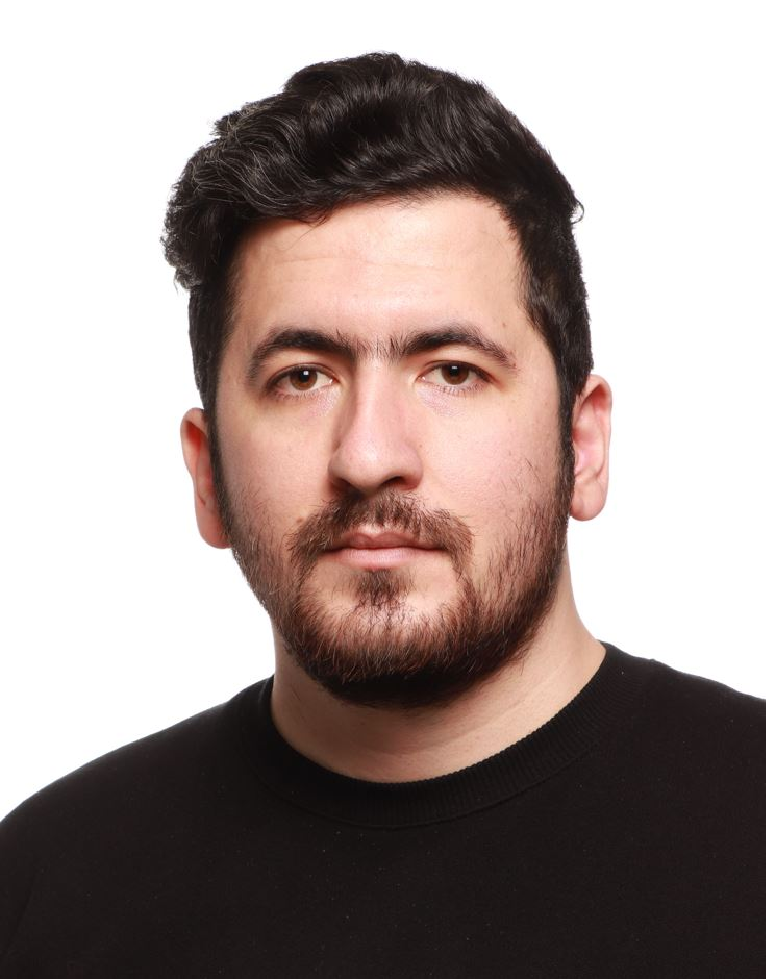}}]{Mahdi Hejrati} received his M.Sc. degree in 2021 from Sharif University of Technology (SUT), Tehran, Iran. He is currently a PhD student at the unit of Automation Technology and Mechanical Engineering, Tampere University, Tampere, Finland. His research interests include nonlinear model-based control, physical human-robot interaction, and human-inspired control.
\end{IEEEbiography}

\begin{IEEEbiography}[{\includegraphics[width=1in,height=1.25in,clip,keepaspectratio]{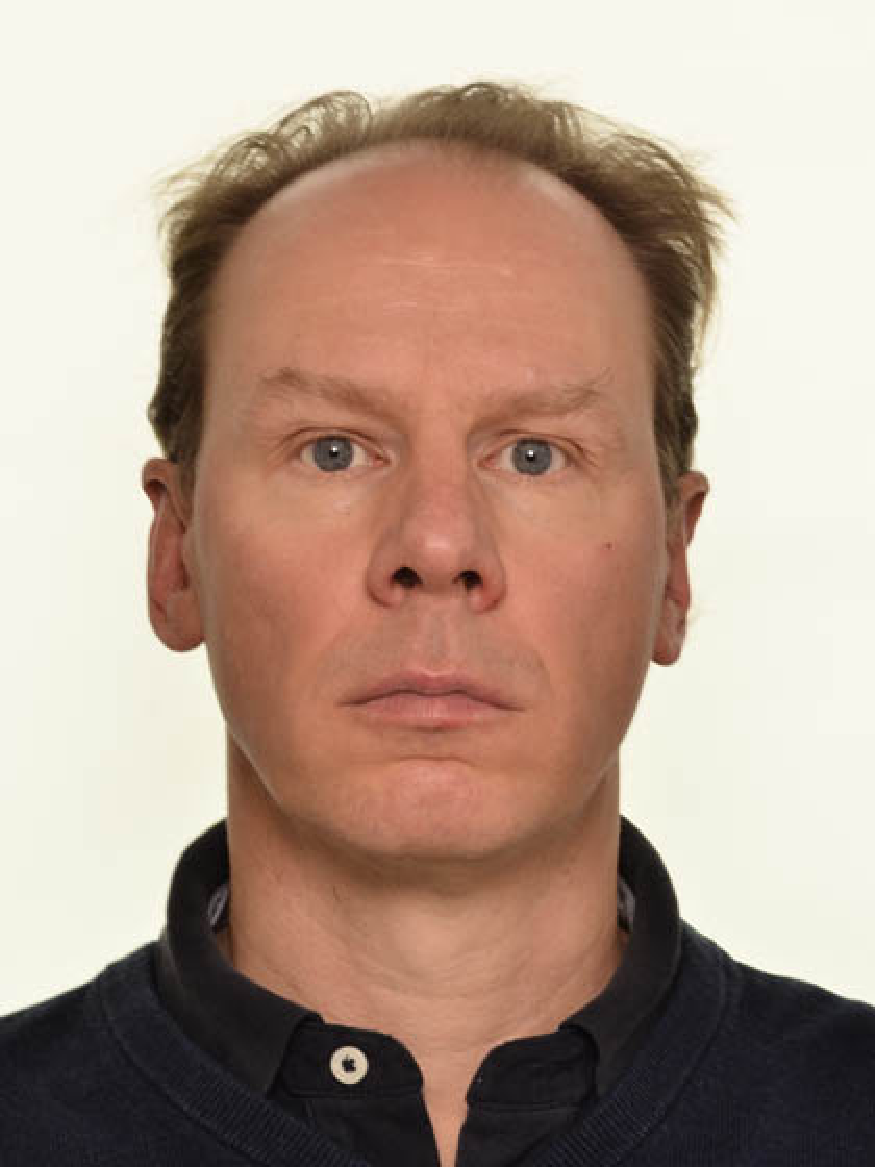}}]{Jouni Mattila}
Dr. Tech. received his M.Sc. (Eng.) in 1995 and Dr. Tech. in 2000, both from Tampere University of Technology (TUT), Tampere, Finland. He is currently a Professor of machine automation with the unit of Automation Technology and Mechanical Engineering, Tampere University. His research interests include machine automation, nonlinear model-based control of robotic manipulators and energy-efficient control of heavy-duty mobile manipulators.
\end{IEEEbiography}

\vfill

\end{document}